%% file: main.tex
\documentclass[lettersize,journal]{IEEEtran}
\usepackage{amsmath,amsfonts}
\usepackage{algorithmic}
\usepackage{array}
\usepackage[caption=false,font=normalsize,labelfont=sf,textfont=sf]{subfig}
\usepackage{textcomp}
\usepackage{stfloats}
\usepackage{url}
\usepackage{verbatim}
\usepackage{graphicx}
\usepackage{amsmath}
\usepackage{algorithm}
\usepackage{algorithmic}

\usepackage{pifont}
\usepackage{subcaption}
\usepackage{graphicx}

\usepackage{multirow}
\usepackage{multicol}
\usepackage{caption}
\usepackage{xcolor,colortbl}
\usepackage{booktabs}       
\usepackage{amsfonts}
\usepackage{graphicx}
\usepackage{amsmath}
\usepackage{multirow}
\usepackage{caption}
 
\usepackage{xcolor,colortbl}

\usepackage{amsmath}
\usepackage{amssymb}
\usepackage{mathtools}

\usepackage{siunitx}
\usepackage{xspace}
\usepackage{lipsum}

\usepackage{adjustbox}

\newcommand{\e}{\epsilon}
\newcommand{\y}{\gamma}

\newcommand{\N}{\mathcal{N}}

\newcommand{\B}{\mathcal{B}}

\DeclareMathOperator*{\argmin}{argmin}
\DeclareMathOperator*{\argmax}{argmax}

\def\BibTeX{{\rm B\kern-.05em{\sc i\kern-.025em b}\kern-.08em
    T\kern-.1667em\lower.7ex\hbox{E}\kern-.125emX}}
\usepackage{balance}
\IEEEoverridecommandlockouts                              

\title{Expert Behavior Prior Reinforcement Learning}

\author{Gong Gao$^{1}$, Weidong Zhao$^{1*}$,
Xianhui Liu$^{1}$, and Ning Jia$^{1}$
\thanks{*This study was supported by National Major Science and Technology Special Project (CN) (Grant Number 2025ZD1604900).}
\thanks{$^{1}$ School of Computer Science, Tongji University, China 
{\tt\small g18438613630@126.com},
{\tt\small weidongzhao111@gmail.com},
{\tt\small xianhui\_l@163.com},
{\tt\small jianing7072@tongji.edu.cn}
}%
}

\begin{document}

\maketitle
\thispagestyle{empty}
\pagestyle{empty}


\input{section/abstract}

\IEEEpeerreviewmaketitle
\input{section/0_Introduction}
\input{section/1_Preliminaries}
\input{section/4_Related_Works}

\input{section/2_Methods}

\input{section/3_Experiments}

\input{section/5_Conclusion}

\small
\bibliography{IEEE}
\bibliographystyle{IEEEtran}


\begin{IEEEbiography}[{\includegraphics[width=1in,height=1.25in,clip,keepaspectratio]{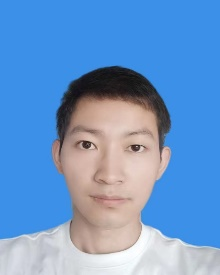}}]{Gong Gao}
received the M.S. degree in computer science from Sichuan University, Chengdu, China. He is currently working toward the Ph.D. degree in computer technology at Tongji University, Shanghai, China. His research interests include computer vision, continual Learning and reinforcement learning. 
\end{IEEEbiography}
\begin{IEEEbiography}[{\includegraphics[width=1in,height=1.25in,clip,keepaspectratio]{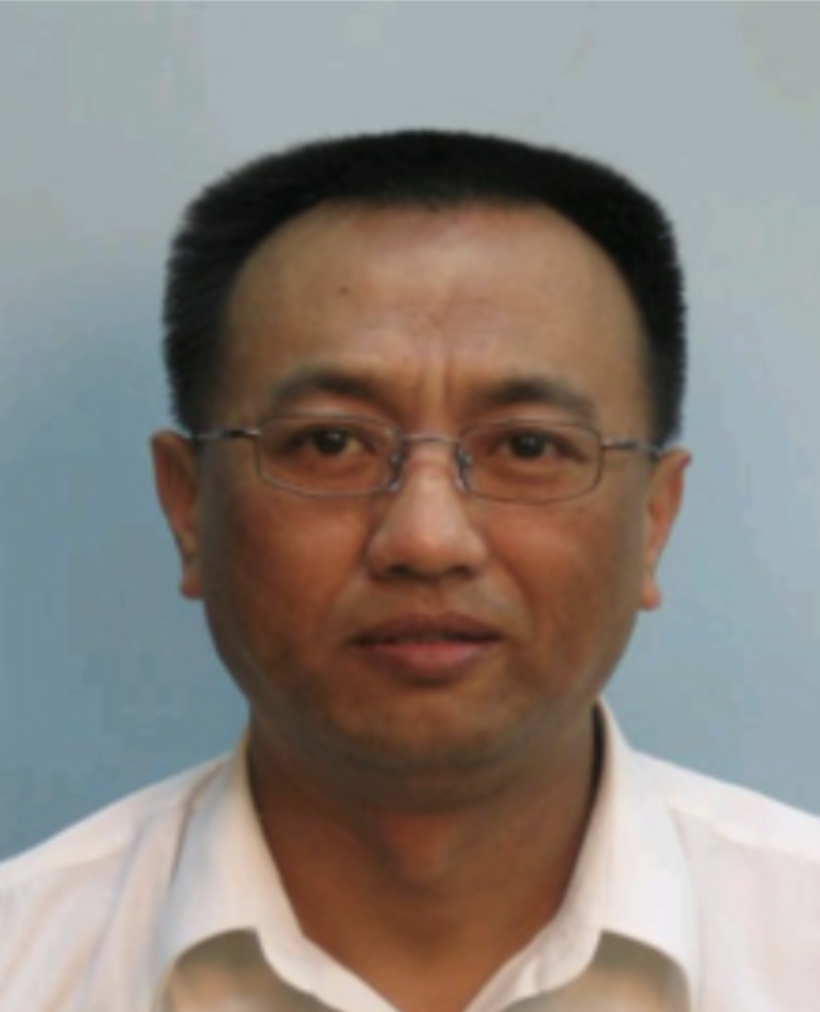}}]{Weidong Zhao}
received the Ph.D. degree in computer science from Tongji University, Shanghai, China, in 2004.
He is currently a Professor with Tongji University. 
Dr. Zhao is a Member of China National Technical Committee for Industrial Automation Systems and Integration Standardization, the Chief Expert on Information Technology in Sci-Tech Engineering of Manufacturing Industry during the eleventh five-year
project of the Ministry of Science and Technology of China, and the Team Leader of Information Technology Manufacturing Engineering of Shanghai Municipal Science and Technology Commission.
\end{IEEEbiography}

\begin{IEEEbiography}[{\includegraphics[width=1in,height=1.25in,clip,keepaspectratio]{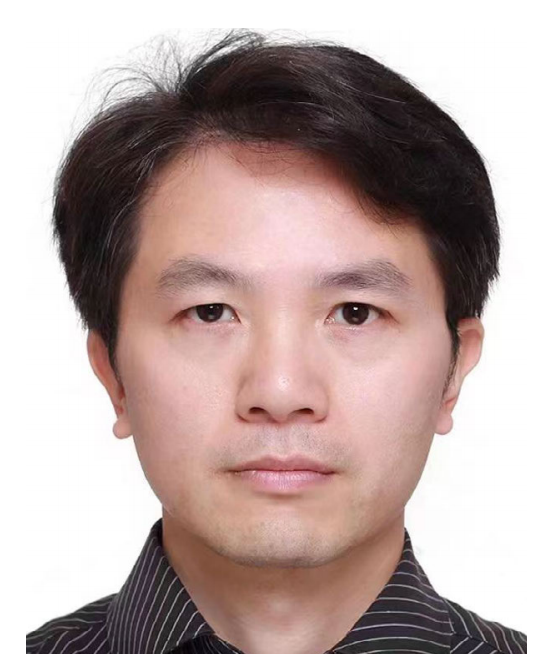}}]{Xianhui Liu}
received the Ph.D. degree from Tongji University, Shanghai, China, in 2014. He is currently an Associate Professor with the College of Electronic and Information Engineering, Tongji University. He is also the Deputy Director of the CAD Research Center, Tongji University. His current research interests include machine learning, data mining, big data, and networked manufacturing. He is a member of the Artificial Intelligence Committee of Shanghai Computer Association.
\end{IEEEbiography}
\begin{IEEEbiography}[{\includegraphics[width=1in,height=1.25in,clip,keepaspectratio]{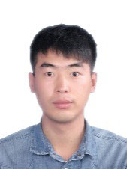}}]{Ning Jia} received the B.S. degree in information management and information systems and the M.S. degree in computer technology from the Shandong University of Science and Technology in 2012 and 2014, respectively, and the Ph.D. degree in computer science and technology from Tongji University, Shanghai, China, in 2019. He is currently a Post-Doctoral Researcher with the College of Electronic and Information Engineering, Tongji University. His current research interests include machine learning.
\end{IEEEbiography}

\normalsize

\input{section/6_Appendix}

\end{document}

%% file: section/abstract.tex
\vspace{-1em}
\begin{abstract}
Behavior prior reinforcement learning (BPRL) has emerged as a promising paradigm to improve sample efficiency in online reinforcement learning (RL) by leveraging policy priors derived from offline demonstrations. However, most existing BPRL methods rely on static offline datasets, which often suffer from low data diversity and suboptimal trajectory quality. This reliance restricts the effectiveness of policy priors, hindering both policy exploitation and stability during online training. Consequently, agents are prone to inefficient exploration and unstable learning dynamics.
To address these limitations, we deviate from existing offline pre-training methods and propose an Expert Behavior Prior (EBP) algorithm. Specifically, we introduce a Q-guided conditional variational autoencoder (Q-CVAE) that learns to generate expert policy priors directly from the online replay buffer. This enables the generation of high-value actions for guiding policy updates without relying on pre-collected expert trajectories. To further enhance policy exploitation, we propose an expert policy guidance (EPG) mechanism that selects expert actions from a generative support set, and we integrate a policy gradient correction (PGC) module to harmonize Q-guidance with expert supervision, promoting stable and consistent policy improvement.
Extensive experiments conducted on robotic control (Gym, PyBullet) and industrial control (DMControl) benchmarks demonstrate that EBP significantly outperforms state-of-the-art online RL algorithms, achieving higher sample efficiency and more stable convergence. 

\end{abstract} 

%% file: section/0_Introduction.tex
 
\section{Introduction}
\label{Introduction}
Online reinforcement learning (RL) has made significant progress in recent years, yet it remains notoriously sample inefficient, particularly when compared to offline paradigms such as imitation learning~\cite{vahabpour2024diverse,agrawal2025markov,jia2025x}.
One potential explanation lies in the reliance on the Bellman equation in most online RL algorithms, which often provides inaccurate value estimates. This can misguide policy updates and exacerbate exploration inefficiency~\cite{quangaugmenting} due to epistemic uncertainty in the Q-function~\cite{ghasemipour2022so,van2025epistemic}. 

To tackle this challenge, behavior prior reinforcement learning (BPRL)~\cite{JMLR:v23:20-1038,yuan2024pre} has emerged as a prominent research direction. As a generalization of standard online RL, BPRL pretrains a behavior cloning model on offline datasets~\cite{guo2024blend} and leverages a supervised learning paradigm to guide policy updates~\cite{uchendu2023jump,wang2024scaling,xia2024learn}.
In existing BPRL algorithms, offline pre-training is critical in accelerating policy convergence. First, offline pre-training leverages a behavior cloning model training on offline datasets to provide an expert policy prior, offering explicit guidance for policy updates and improving sample efficiency~\cite{ball2023efficient,wagenmaker2023leveraging}.
Second, offline pre-training helps mitigate ineffective exploration driven by Q-guidance, thereby enabling more efficient policy improvement~\cite{daoudi2024enhancing}. This process involves incorporating gradients from both Q-guidance and expert behavior priors, followed by the construction of an appropriate weighting mechanism to ensure stable policy improvement.

However, current BPRL methods suffer from a critical limitation: the policy prior is constrained to samples drawn from static offline datasets, thereby restricting policy quality to that of the pre-collected trajectories~\cite{hao2023leveraging,xudong2024iterative}. In practical scenarios, obtaining high-value trajectory data is often infeasible due to the substantial cost and safety concerns associated with data collection~\cite{chemingui2025constraint}.
This limitation in data quality presents two major challenges. 
First, it hinders the generation of high-value action priors, which could otherwise facilitate stable policy updates for the agent.
Second, it impairs the ability of the policy to provide informative gradients for the agent, thereby weakening its capacity to correct suboptimal gradient directions induced by Q-guidance.
To this end, we propose an alternative approach—rather than relying on offline expert trajectory data, we draw upon advances in generative modeling to actively generate policy priors. While prior works have explored using VAEs to reconstruct offline data for rapid policy transfer~\cite{rana2023residual,wilcoxson2024leveraging}, leveraging generative models to enhance policy exploitation in online RL remains underexplored, to the best of our knowledge.

Therefore, in this paper, we seek to answer a fundamental question: Can existing generative reconstruction methods be directly applied to incorporate policy priors during online RL learning, without relying on offline expert trajectory pretraining or explicit datasets constraint~\cite{ran2023policy,li2023accelerating,lyu2025cross}, to achieve stable policy improvement?
To explore this, we employ a Q-guided loss function during the online learning process to train a behavior cloning model for reconstructing high-value actions, while utilizing its generative expert policy priors to enhance the policy update process. We conduct experiments with a fixed set of parameters across three challenging continuous control environments. The experimental results demonstrate that EBP consistently achieves significant performance improvements. 
Furthermore, we demonstrate that EBP can maintain stable policy updates even when trained in noisy environments.
Finally, the effectiveness of each module of EBP is demonstrated through ablation experiments.
These results provide strong empirical evidence that EBP offers a stable, efficient, and scalable solution for training online RL agents.

The main contributions of this paper are as follows.
\begin{itemize}
\item We propose an EBP algorithm that learns policy priors from the online replay buffer and generates high-value actions, thereby guiding the policy update process in online RL.

\item We propose integrating Q-guidance and EPG into the policy update process, leveraging a supervised learning paradigm for policy optimization. This approach effectively mitigates the inefficient exploration challenges commonly encountered in online RL algorithms.

\item We introduce a PGC mechanism that leverages EPG gradients to rectify the Q-guidance gradients, thereby enabling stable policy updates. Additionally, we observe that despite the presence of reward noise during the training phase, the performance gains remain stable.

\item We conduct extensive empirical experiments and provide detailed analysis.
Performance evaluations on two robotic control and one industrial control benchmarks consistently demonstrate the high sample efficiency and stability of the proposed method.
 
\end{itemize}

The remainder of this paper is organized as follows. Section~\ref{sec:pre} introduces the background. Related work is reviewed in Section~\ref{related_wor}. In Section~\ref{sec:method}, we present the proposed Expert Behavior Prior (EBP), including the overall framework and convergence analysis. Section~\ref{sec:experiment} reports experimental results on three continuous control environments, demonstrating the effectiveness and practicality of EBP. We discuss the contributions and limitations of our approach in Section~\ref{appendix_discussion}. Finally, Section~\ref{sec:con_f} concludes the paper and outlines potential future directions.

%% file: section/1_Preliminaries.tex
 
\section{Background}
\label{sec:pre}
\subsection{Value-based RL}
In reinforcement learning, the standard framework is typically modeled as a Markov Decision Process (MDP) defined by the tuple $(S, A, T, R, \gamma)$~\cite{kumar2023policy}, where $S$ is the state space, $A$ is the action space, and $\gamma \in [0,1)$ is the discount factor. The environment is characterized by two attributes: the transition function $T:S \times {A} \to {S}$, and the reward function $R:S \times {A} \to {R}$. The objective of an RL agent is to learn a policy $\pi : S \to A$, which maximizes the expected cumulative discounted return: $J(\pi) = {\mathbb{E}_{s_t}} [\sum\nolimits_{t=0}^{\infty} {\gamma^t r_t \mid a_t = \pi(s_t)} ]$, where ${\mathbb{E}}$ denotes the expectation.

A typical online RL algorithm can be defined by integrating two optimization processes: the Critic and the Actor. The first involves training the Q-network using temporal difference error, which can be formulated as
\begin{equation} 
\label{eq:belleman}
\theta^{*} ={\mathop {\argmin}\limits_{\theta}}\ \mathbb{E}_{s,a,r,s^{'},a^{'}}{[Q_\theta(s,a )-y]^2}, \\
\end{equation}
where $y=r + \y \max_{a^{'}} Q_{\theta'}({s^{'}}, a^{'})$ is the Bellman optimal Q-value, $s, a, r, s^{'}, a^{'}$ is state, action, immediate reward, next state, and next action, respectively.
The second part involves training the Actor network, which can be formulated as
\begin{equation}
{\phi^{*}} = {\mathop {\argmin}\limits_{\phi}}\ -{\mathbb{E}_{s}}[{Q_\theta }(s,\pi_{\phi}(s))].
\end{equation}

In the policy improvement step, the objective function varies across RL algorithms but consistently depends on estimated value functions. The greedy optimization of this objective can lead to suboptimal policies, particularly when value estimates are inaccurate during initial training~\cite{sutton1998reinforcement,jin2018q,quangaugmenting}. To refine these estimates, the agent needs to explore the environment and collect more relevant data. Given limited prior knowledge, such exploration is typically guided by introducing stochastic noise into the policy. While this can help reveal high-value regions, it may also direct the agent towards uninformative areas~\cite{daoudi2024enhancing}. More importantly, during the training phase, such random exploration can lead to undesirable policy oscillations, compromising the stability of the learning process.

\subsection{Policy Prior Guidance for Online RL}
Policy priors, which represent behavioral references derived from offline data, can further guide policy learning.
Due to its favorable applicability and effectiveness, this approach has been widely adopted in offline-to-online RL settings~\cite{yu2023actor, zheng2023adaptive, luo2025optimistic}. 
In these approaches, a policy prior $ G_\omega({s})$ is applied to ensure that the current output ${\pi _\phi }$ remains proximate, which can be formulated as
\begin{equation}
 ||({\pi _\phi }(s), G_\omega({s }) ) ||_2 \le \varepsilon_a,
\end{equation}
where $||\cdot||_2$ indicates Euclidean distance, $\varepsilon_a $ denotes a constant value that the divergence between the current policy ${\pi _\phi }$ and the policy prior $G_\omega$ is bounded, $\phi$ and $\omega$ represent the parameters of the agent and the behavior policy prior, respectively.
This method accelerates policy convergence by distilling expert prior knowledge into the agent and proves effective when learning from expert offline data. However, BPRL is limited by the quality of offline data and cannot provide high-value prior guidance during policy updates, often leading to low sample efficiency and suboptimal performance.

%% file: section/4_Related_Works.tex
\section{Related Works}
\label{related_wor}

\subsection{Offline Pre-training for Online RL} 
We highlight the interconnections between online RL and the transition from offline-to-online settings. A significant portion of prior works have involved performing offline RL~\cite{batchRL,lyu2022mildly,offlineRL,huang2024pessimism} followed by online fine-tuning~\cite{nakamoto2023cal,awac,ball2023efficient,wagenmaker2023leveraging}. 
Notably, Cal-QL~\cite{nakamoto2023cal} learned a calibrated value function that facilitates effective online fine-tuning, allowing the method to leverage the advantages of offline pre-training during the online adaptation phase. 
However, the issue of data distribution shift in offline-to-online algorithms remains unresolved.
In contrast, our proposed algorithm introduces a generative policy prior to guide the agent update, enhancing online learning performance without the need for offline pretraining and significantly reducing system complexity.

\subsection{Online RL with Offline Policy Priors}
In contrast to training agents directly from scratch, RL with offline policy prior~\cite{walke2023don,ball2023efficient,zang2023behavior} leverages explicit knowledge obtained from offline datasets to guide online agent training, thereby accelerating policy convergence. In practical applications, these additional priors are typically used to direct the agent's focus on high-value areas of the MDP, reducing unnecessary exploration. Initially, these data are employed to initialize policies through behavior cloning (BC)~\cite{george2023one}, and they permeate throughout the optimization process. For example, RLPD~\cite{ball2023efficient} and EXPLORE~\cite{li2023accelerating} incorporated demonstration data into the replay buffer and introduced additional regularization terms to encourage the optimization process to better exploit the available demonstrations. Similarly, POfD~\cite{kang2018policy}, LOGO~\cite{rengarajan2022reinforcement}, and BPR~\cite{zang2023behavior} imposed penalties or constraints on the reinforcement learning objective, forcing the agent's policy to remain close to the policy prior.
However, these methods are only feasible when expert policies are available, as they cannot effectively guide policy updates in states with low-quality interactions~\cite{hao2023leveraging,xudong2024iterative,xudong2024goal}. Additionally, these methods often require large amounts of pretraining data to be effective, which exacerbates their lack of scalability and limits their deployment across diverse environments~\cite{rengarajan2022reinforcement}.

\subsection{Online RL with Online Policy Priors}
More recent work has attempted to move beyond static priors by constructing online policy priors during training. These methods aim to provide flexible and adaptive guidance by leveraging either historical trajectories~\cite{kapturowski2018recurrent,quangaugmenting}, or local policy approximations~\cite{zhao2023learning,shen2021theoretically,guo2024blend,daoudi2024enhancing}.
On one hand, incorporating historical interactions is an intuitive strategy to enhance the decision-making process.
For instance, R2D2~\cite{kapturowski2018recurrent} employed LSTM~\cite{yu2019review}-based recurrent memories to improve temporal context. Similarly, ALH~\cite{quangaugmenting} improved decision-making by using historical observations to guide the current agent’s actions, leading to notable performance gains.
However, the incorporation of additional temporal state representations can lead to higher similarity among generated actions, which may in turn degrade overall performance, as observed in PEER~\cite{he2023frustratingly}.

On the other hand, nearest-neighbor reinforcement learning~\cite{shah2018q} algorithms implement constrained policy exploration and reduce ineffective exploration due to Q-value estimation bootstrapping.
RLLG~\cite{daoudi2024enhancing} proposed an RL algorithm based on noisy policy switching, which effectively leverages local guide policies to significantly enhance the performance of Approximate Policy Evaluation (APE)-based RL algorithm, with particularly notable improvements during the early stages of training.
IRA~\cite{gao2026improving} retrieves nearest-neighbor high-quality trajectories from historical interactions to guide policy updates, leading to significant improvements in sample efficiency.
Our approach shares conceptual similarities with existing works~\cite{shen2021theoretically,daoudi2024enhancing,gao2026improving}, as it also incorporates local guidance to acquire policy prior knowledge for policy updates.
The key distinction lies in our use of a Q-guided CVAE to generate expert policy priors, enabling an effective integration of expert-driven supervision and Q-guided optimization during policy learning.

Several recent works~\cite{ji2023seizing,luo2024offline} seek to better exploit high-quality samples in the replay buffer by incorporating historically high-value actions into Q-value updates. BAC~\cite{ji2023seizing} addresses Q-value underestimation in later training stages by gradually shifting from action-value to state-value estimation. OBAC~\cite{luo2024offline} concurrently trains an offline policy using the shared online replay buffer and employs value-based comparisons to identify an outperforming offline policy, which is then used as an adaptive constraint for online policy updates. However, both approaches increasingly rely on historical high-value actions in the later stages of training, which may bias policy updates toward conservative behaviors and weaken exploration, particularly in environments with complex dynamics or multi-modal reward structures.

%% file: section/2_Methods.tex
\section{Methods}
\label{sec:method}

\begin{figure}[ht]
\begin{center}
\centerline{\includegraphics[width=0.5\textwidth]{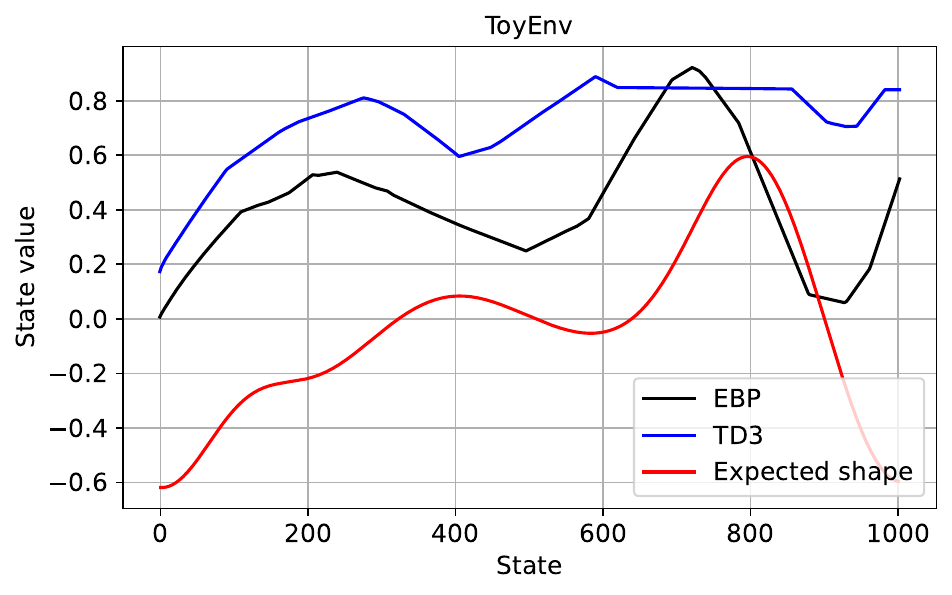}}
\caption{Detailed description of the Toy environment and a comparative analysis of the policies learned by the EBP and TD3 algorithms.}
\label{fig:toy_results}
\end{center}
\vskip -0.2in
\end{figure}
To better monitor the value-based RL agents, we introduce
a simple simulation environment named ToyEnv:
$S$ = [0, 1000]; $A$=[-6, 6]; $T(s, a)$ = $s$ + $a$. Given $\Theta $ =
\{(0, -81),(1, -27),(2, 12),(3, -9),(4, 81),(5, -81)\}, the
reward for every tuple of $(s, a)$ relies on the next state:
\begin{equation}
\begin{aligned} 
    u(x,\beta,\sigma)&={\frac{1}{\sigma \sqrt {5\pi } } }{e^{ - 0.1{{(\frac{{x - \beta}}{\sigma})}^2}}}\\
    R(s,a)&=\sum\limits_{(\beta ,\sigma) \in \Theta } {u(T(s,a),\beta ,\sigma)} .\\
\end{aligned}
\end{equation}

We conducted comparative experiments using the proposed EBP algorithm and the vanilla TD3 algorithm in the ToyEnv environment, with each method trained for $2$M timesteps. As a diagnostic baseline, we visualize the learned state value distributions over the one-dimensional state space in ToyEnv.
After training, we evaluated the predicted state values of both EBP and TD3. Specifically, the state value is computed as $Q_{\theta}(s, \pi(s))$, which represents the action value induced by the current policy. The resulting state value profiles over the entire one-dimensional state space are illustrated in Fig.~\ref{fig:toy_results}.

As shown in Fig.~\ref{fig:toy_results}, EBP more accurately captures the overall structure of the state value distribution. In particular, within the state range $S \in [0, 900]$, EBP can model the reward trend more faithfully. Moreover, compared to TD3, EBP produces more conservative value estimates that are closer to the ground truth, indicating that the prior-guided policy updates effectively mitigate the overestimation bias commonly observed in TD3. However, in the region $S \in [900, 1000]$, neither EBP nor TD3 can learn accurate state values. We attribute this failure to insufficient exploration: since the highest rewards are attained in the range $S= 800$, the learned policies cease exploring further to the right, leading to sparse sampling and consequently inaccurate Q-value estimation in that region.

\begin{figure}[ht]
\begin{center}
\centerline{\includegraphics[width=0.5\textwidth]{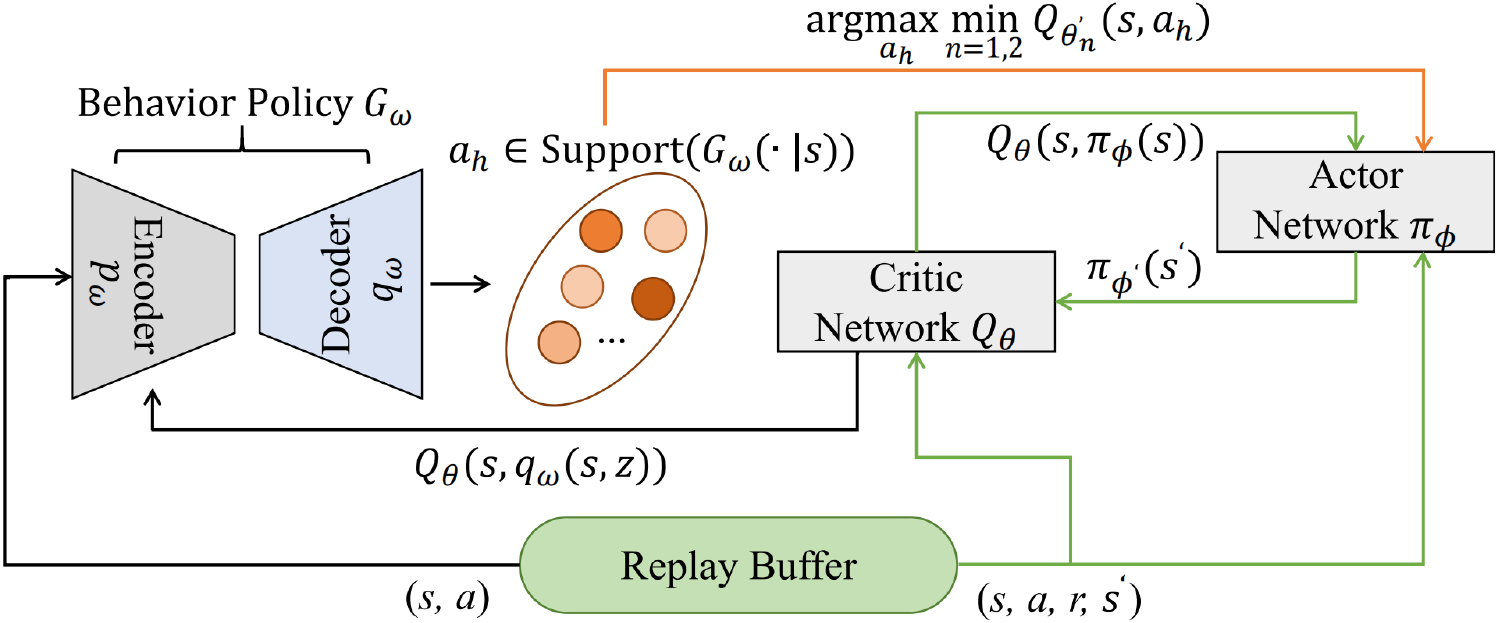}}
\caption{Overview of the proposed Expert Behavior Prior (EBP) framework, including both the Q-CVAE training stage and the RL training process. The action-value function (Q) is first used to guide the behavior policy in generating multiple high-value support set actions, from which an expert action is selected via Q-value evaluation and subsequently used to guide the policy update.}
\label{fig:block_diag}
\end{center}
\vskip -0.2in
\end{figure}

In combination with actor-critic architecture, we have devised the EBP algorithm, as shown in Fig.~\ref{fig:block_diag}.
First, we propose Q-CVAE to learn policy priors from the online replay buffer and dynamically generate high-value actions during the RL training process. 
Furthermore, we introduce the EPG module that selects expert actions as anchors to guide the policy update process, thus enhancing the sample efficiency of online RL. To further improve stability, we propose a gradient correction mechanism to balance the influence of Q-guidance and expert prior guidance. The complete training procedure is summarized in Algorithm~\ref{alg:EBP_online}.

\subsection{Online Training Policy Prior}
In this section, we provide a comprehensive overview of the construction process underlying the CVAE model. Given the mode collapse issue in GANs, we avoid using GAN-based networks to learn policy priors. Instead, we employ a CVAE~\cite{fujimoto2019off} to construct the behavior policy model $G$, which leverages a Q-guided loss to learn high-value action reconstruction. Previous studies~\cite{kingma2013auto} have shown that latent-variable models possess strong expressive power and can approximate complex data distributions given sufficient model capacity. Building upon this property,~\cite{huang2023reparameterized} further employed CVAE-based policy parameterization to model complex action distributions in reinforcement learning. However, directly learning such distributions under non-stationary optimization dynamics remains challenging. Motivated by this observation, we argue that a CVAE trained using samples from the current replay buffer is sufficient to provide an effective policy prior for guiding actor optimization.
The hyperparameter configuration for the Q-CVAE is outlined in Section~\ref{sec:alg_detaild}.


We first train the Q-network to establish a foundation for guiding the CVAE in generating high-value action, while also enabling the effective evaluation of the support set actions produced by the Q-CVAE.
The loss of the Q-network is the temporal difference error, which can be formulated as
\begin{equation}
\label{eq:critics}
\centering
   {{L}_Q}(\theta ) = {[Q_{\theta}(s,a ) - y]^2},
\end{equation}
where $y=R(s,a) + \y \min_{n=1,2} Q_{\theta'_n}(s',{\pi _{\phi^{'}} }(s^{'}))$.
Sampling from the replay buffer, the goal of the CVAE is to reconstruct actions conditioned on states, such that the reconstructed actions follow the same distribution as those in the training data.
The CVAE model $G_\omega(s)$, parameterized by $\omega$, consists of an encoder $p_\omega(s,a)$ and a decoder $q_\omega(s,z)$, denoted as $G = \{p, q\}$. The CVAE is optimized by maximizing its variational lower bound, which is equivalent to minimizing the following objective function, which can be formulated as
\begin{equation}
    \label{eq:Rec}
    \begin{aligned}
    \mathcal{H}_{\text{Rec}}(\omega)=& \mathbb{E}_{(s,a)\sim\mathcal{\B},z\sim p_\omega(s,a)}[ (a - q_\omega(s,z))^2 \\
    &+ {\text{KL}}(p_\omega(s,a), \mathcal{N}(0,{\bf I})) ],
    \end{aligned}
\end{equation}
where ${\text{KL}}(p,q)$ denotes the KL divergence between the probability distributions $p(\cdot)$ and $q(\cdot)$, and $\mathbf{I}$ represents the identity matrix. When sampling actions from the CVAE, we first sample a latent variable $z$ from the prior distribution, which is assumed to follow a normal distribution $\mathcal{N}(0, \mathbf{I})$. This latent variable, along with the state $s$, is then passed into the decoder $q_\omega(s,z)$ to obtain the desired decoded action.

Since Eq.~\eqref{eq:Rec} only reconstructs actions consistent with the training data distribution, the actions generated by the CVAE model essentially correspond to a simple behavior cloning, failing to fully leverage the generative capabilities of the CVAE model.
As noted by DIDI~\cite{liu2024didi}, employing Q-value guided gradients to train a generative model yields diverse and high-value actions. Inspired by~\cite{liu2024didi}, we propose a Q-guided loss that employs a double Q-network to guide the gradient update direction of the CVAE, and the Q-value guided loss can be formulated as
\begin{equation}
    \label{eq:cvae_q}
     \mathcal{H}_{\text{Q}}(\omega) = - 0.5  \sum\nolimits_{{\textit{n}} = 1,2}{Q_{\theta_n}(s,q_\omega(s,z))}  .
\end{equation}

The complete Q-CVAE model is optimized by integrating these two loss components, and the overall loss function can be formulated as
\begin{equation}
    \label{eq:CVAE}
    \mathcal{H}_{G}(\omega) = \mathcal{H}_{\text{Rec}}(\omega) +\alpha \mathcal{H}_{\text{Q}}(\omega),
\end{equation}
where $\alpha$ denotes the weighting parameter of loss $\mathcal{H}_{\text{Q}}$. In the experiments detailed in Section~\ref{sec_abla}, we employ multiple sets of experiments to demonstrate the effect of the parameter $\alpha$ on the improvement of policy quality.

\subsection{Expert Policy Guidance}
\label{sec:hpg}
Unlike the Actor network optimization process in TD3~\cite{fujimoto2018addressing}, we guide the Actor network update using the average of two Q-networks, which can be formulated as
\begin{equation}
\label{eq:pi}
\begin{aligned}
J_{Q}(\phi)= -0.5   {\mathbb{E}_{s}}\left[ \sum\nolimits_{{\textit{n}} = 1,2}{Q_{\theta_n} }(s,{\pi _\phi }(s)) \right].
\end{aligned}
\end{equation}

We utilize the behavior policy model $G$ that generates the policy prior in $G_\omega(s)$, which is essential for providing the optimal anchor for the Actor update process. 
The behavior prior is obtained through multiple inferences of the generative network, forming a policy support set from which the action with the highest value is selected. This process can be formulated as
\begin{equation}
    \label{eq:optimal_a}
    {\tilde {a}} = {\mathop{\argmax}\limits_{a_h}} (\min_{n=1,2}(Q_{\theta_{n}^{'}}(a_h))), a_h \in \text{Support}(G_\omega(\cdot|s)),
\end{equation}
where $\mathop{\argmax}\limits_{a_{h}}$ denotes the process that identifies expert action from the policy prior set $\text{Support}(G_\omega(\cdot|s))$, and $h = [1, 2, 3, \dots, H]$ denotes the indices of $H$ prior actions generated by the Q-CVAE.
Following the identification of the expert action, we implement a systematic optimization mechanism that progressively refines the policy network with the expert action through a supervised learning paradigm consistent with the algorithm established in~\cite{guo2024blend}.
By treating the constraint term as a penalty, we minimize the following objective
\begin{equation}
\label{eq:KLreg}
{\mathop{\argmin}\limits_\phi}\  \mathbb{E}_{s} [ || \pi_\phi(s)-\tilde a||_2^2].
\end{equation}

With expert action $\tilde a$ defined, Eq.~\eqref{eq:KLreg} is equivalent to the following policy training objective
\begin{equation}
\label{eq:epg}
\begin{aligned}
J_{\text{Sup}}(\phi)={\mathbb{E}_{s}}\left[ {{({\pi _\phi }(s) - {{\tilde a}})}^2} \right].
\end{aligned}
\end{equation}

By Eq.~\eqref{eq:epg}, ${\pi_\phi}$ explores within this constrained exploration region, thereby facilitating gradual policy improvement.
The overall loss function of the Actor network can be formulated as
\begin{equation}
J_{\pi}(\phi)=   J_{\text{Q}}(\phi)+ \mu J_{\text{Sup}}(\phi),
\end{equation}
where $\mu$ denotes the weight of the policy guidance loss $J_{\text{Sup}}$.  
Ideally, expert policy prior guidance should correct the gradient of Q-guided reinforcement learning, reduce inefficient exploration, and improve sample efficiency.
Although the Actor network update process enhances sample efficiency by combining Q-guidance with EPG, this bi-objective optimization may introduce potential policy oscillations.  
Specifically, when the gradients of Q-value guidance and EPG exhibit high similarity, excessive update magnitudes can lead to policy nonstationarity.
To this end, we propose the PGC method, which enables support set expert guidance to consistently influence policy updates while progressively enhancing Q-guided exploration, thereby leading to stable performance improvement.

\subsection{Policy Gradient Correction}
During the Actor training phase, the expert action anchor inferred from Q-CVAE is used to compute the loss, resulting in two losses: the Q-guided loss \(J_{\text{Q}}\) and expert policy-guided loss \(J_{\text{Sup}}\).
We aim to align the gradient directions of both losses, ensuring that the policy update directions guided by Q-learning and the expert policy are consistent. When these update directions diverge, our proposed PGC increases the weight of the EPG to correct the Actor update direction. Conversely, when the gradient directions are highly aligned, we employ gradient clipping to prevent excessively large updates of the Actor network.

As defined in Eq.~\eqref{eq:margin_cos}, the weighted term  \( g \) measures the similarity gap between the corresponding gradients \( \nabla J_{\text{Q}} \) and \( \nabla J_{\text{Sup}} \). For each pair of gradients, we compute the cosine similarity \(\frac{\nabla J_{\text{Q}} \cdot \nabla J_{\text{Sup}}}{\|\nabla J_{\text{Q}}\| \|\nabla J_{\text{Sup}}\|} \) and subtract the margin \(m \). The \(\text{ReLU}\) function ensures that only pairs of lines with a similarity lower than \( 1-m \) have their weights computed, thus focusing the gradient alignment on the less similar gradients. The final value is the average of all positions in the gradient grid, which can be formulated as 
\begin{equation}
\label{eq:margin_cos}
g = \frac{1}{\mathbf{H} \times \mathbf{W}} \sum_{i=1}^{\mathbf{H}} \sum_{j=1}^{\mathbf{W}} \text{ReLU} \left( 1 - m - \frac{\nabla J_{\text{Q}} \cdot \nabla J_{\text{Sup}}}{\|\nabla J_{\text{Q}}\| \|\nabla J_{\text{Sup}}\|} \right),
\end{equation}
where both \( \nabla J_{\text{Q}} \) and \( \nabla J_{\text{Sup}} \) are both computed using the weights from the final fully connected layer of the Actor network, $\mathbf{H}$ and $\mathbf{W}$ denotes the size of the gradient map.
Then we get EPG loss with adaptive weighting $g$ as Eq.~\eqref{eq:margin_cos}. The overall loss function of the Actor network can be formulated as
\begin{equation}
\label{eq:actor}
J_{\pi}(\phi)=   J_{\text{Q}}(\phi)+ \mu gJ_{\text{Sup}}(\phi),
\end{equation}
where $\mu$ represents the weighting parameter of the expert guidance loss $J_{\text{Sup}}$. By adjusting $\mu$, we can effectively control the contribution of the loss $J_{\text{Sup}}$, thus achieving an enhanced alignment with our desired optimization objectives.
In our following experiments, we will evaluate the significant role of the \( J_{\text{Sup}} \) loss in achieving the policy improvement for EPG.

\subsection{Actor Network Convergence Analysis}
This section presents a formal convergence analysis of actor updates under combined Q-guidance and expert prior supervision. The actor objective consists of a Q-function guidance term and a supervised loss induced by the expert prior.
\begin{equation}
\begin{aligned}
J_Q(\phi) &= - \mathbb{E}_{s} \left[ Q_\theta(s, \pi_\phi(s)) \right] \\
J_{\text{Sup}}(\phi) &= \mathbb{E}_{s} \left[ \| \pi_\phi(s) - \tilde a \|_2^2 \right].
\end{aligned}
\end{equation}

Then the total loss is:
\begin{equation}
J_{\pi}(\phi) = J_Q(\phi) + \mu g \cdot J_{\text{Sup}}(\phi).
\end{equation}

The gradient of $J_{\pi}(\phi)$ is:

\begin{equation}
\label{eq:full_gradient}
\begin{aligned}
\nabla J_{\pi}(\phi)
&= \nabla J_Q(\phi) + \mu g \cdot \nabla J_{\text{Sup}}(\phi) \\
&= - \mathbb{E}_s \left[ \nabla_a Q_\theta(s, a) \big|_{a=\pi_\phi(s)} \cdot \nabla \pi_\phi(s) \right] \\
&+ \mu g\cdot \mathbb{E}_s \left[ 2 (\pi_\phi(s) - \tilde a) \cdot \nabla \pi_\phi(s) \right] \cdot \nabla J_{\text{Sup}}(\phi) \\
&=\nabla {J_{\text{Q}}+2\mu{ \text{ReLU} \left( 1 - m - \frac{\nabla J_{\text{Q}} \cdot \nabla J_{\text{Sup}}}{\|\nabla J_{\text{Q}}\| \|\nabla J_{\text{Sup}}\|} \right)}}\\
&\cdot \nabla J_{\text{Sup}}(\phi) \\
&=\nabla {J_{\text{Q}}+2\mu \text{ReLU}(1-m - \text{cos}(\nabla J_{\text{Q}},\nabla J_{\text{Sup}}))}\\
&\cdot \nabla J_{\text{Sup}}(\phi) \\
&= \nabla_Q + 2\mu \text{ReLU}(1 - m -\cos(\nabla _{\text{Q}}, \nabla _{\text{Sup}})) \cdot \nabla_{\text{Sup}},
\end{aligned}
\end{equation}
where we define:
$\nabla_Q := \nabla J_{\text{Q}}(\phi)$, $\nabla_{\text{Sup}} := \nabla J_{\text{Sup}}(\phi)$,
$2\mu \text{ReLU}(1 - m - \cos(\nabla_Q, \nabla_{\text{Sup}})) \in [0, 2(2-m)\mu]$.
Specifically, when $\cos(\nabla_Q, \nabla_{\text{Sup}}) \ge 1 - m$, the supervision gradient $\nabla_{\text{Sup}}$ is excluded from the loss computation. This indicates that when the update directions of the two gradients are sufficiently aligned (exceeding the threshold $1 - m$), the influence of $\nabla_{\text{Sup}}$ can be suppressed to stabilize the overall gradient update. Conversely, when a larger deviation between the gradient directions occurs, the weight of $\nabla_{\text{Sup}}$ is accordingly increased to correct the Q-guided gradient $\nabla_Q$.

The gradients of the Q-guided and supervised losses are assumed to be individually bounded; that is, there exist constants $\lambda_1, \lambda_2 > 0$ such that:
\begin{equation}
\label{eq:bounded_gradients}
\begin{aligned}
\| \nabla J_{\text{Q}}(\phi) \| &\leq \lambda_1 \\
\| \nabla J_{\text{Sup}}(\phi) \| &\leq \lambda_2,
\end{aligned}
\end{equation}
where $\nabla J_{\text{Q}}$ denotes the Q-value-based objective, while $\nabla J_{\text{Sup}}$ represents the supervised loss guided by the expert policy prior. Since $\nabla J_{\text{Sup}}$ is computed from a nearest-neighbor anchor set (policy support set), we assume:

\begin{equation}
\lambda_2 \ll \lambda_1.
\end{equation}

An upper bound on the full gradient norm is derived via the triangle inequality and the Cauchy–Schwarz inequality~\cite{horn2012matrix}.
Consequently, Eq.~\ref{eq:full_gradient} is eventually simplified to:
\begin{equation}
\label{eq:gradient_bound}
\begin{aligned}
\| \nabla  &J_{\pi}(\phi)\| \\ 
&= \| \nabla_Q +2\mu \text{ReLU}(1-m-\cos(\nabla_Q, \nabla_{\text{Sup}}))\cdot \nabla_{\text{Sup}} \| \\
&\leq \| \nabla_Q \| +2 \mu\| \text{ReLU}(1 - m - \cos(\nabla_Q, \nabla_{\text{Sup}}))\cdot \nabla_{\text{Sup}} \| \\
&\leq \lambda_1 + 2(2-m)\mu \cdot \| \nabla_{\text{Sup}} \| \\
&\leq \lambda_1 + (4-2m)\mu \lambda_2.
\end{aligned}
\end{equation}

 

Consequently, the gradient norm is upper bounded by $\lambda_1 + (4 - 2m)\mu \lambda_2$. 
Similarly, the gradient norm \( \| \nabla J_{\pi}(\phi) \| \) is lower bounded by \( \lambda_1 \). Therefore, we obtain the following bound:
\begin{equation}
\lambda_1  \le \| \nabla J_{\pi}(\phi) \| \le \lambda_1 + (4 - 2m)\mu \lambda_2.
\end{equation}

Given that $\lambda_2 \ll \lambda_1$, the influence of the expert supervision remains limited, introducing only minor perturbations while enhancing update stability through improved gradient alignment.

\begin{algorithm}[ht]
   \caption{Expert Behavior Prior}
   \label{alg:EBP_online}
\begin{algorithmic}
   \STATE Initialize behavior policy network $G_{\omega}$, critic networks $Q_{\theta_{1}}$, $Q_{\theta_{2}}$, and actor-network $\pi_\phi$ with random parameters $\omega$, $\theta_{1}$, $\theta_{2}$, $\phi$
   \STATE Initialize target networks $\theta'_1 \leftarrow \theta_1$, $\theta'_2 \leftarrow \theta_2$, $\phi' \leftarrow \phi$
  \STATE Initialize empty replay buffer $\B$
     \WHILE{True}
   \STATE Select action with exploration noise $a \sim \pi_\phi(s) + \e$,
   \STATE  observe reward $r$, and new state $s'$
   \STATE Store transition tuple $(s, a, r, s')$ in $\B$
      \STATE Sample mini-batch of transitions $(s, a, r, s')$ from $\B$
        \STATE Train critics:
   \STATE Update critic by minimizing Eq.~\eqref{eq:critics}
    \STATE Train behavior policy:
   \STATE Update Q-CVAE by minimizing Eq.~\eqref{eq:CVAE}
    \STATE Train actor:
     \STATE Get the expert action anchor $\tilde a$ by Eq.~\eqref{eq:optimal_a}
   \STATE Update actor by minimize Eq.~\eqref{eq:actor}
   \STATE $\theta'_n \leftarrow \tau \theta{_n} + (1 - \tau) \theta'_n$
   \STATE $\phi' \leftarrow \tau \phi + (1 - \tau) \phi'$
   \ENDWHILE
\end{algorithmic}

\end{algorithm}

%% file: section/3_Experiments.tex
\section{Experiments}
\label{sec:experiment}
In this section, we employ eight Gym continuous tasks, four PyBullet continuous tasks, and eight DeepMind Control (DMControl) suit tasks to evaluate the performance of the EBP, aiming to answer the following questions: 
(1) Can EBP leverage behavior policy prior to accelerate online learning?
(2) Can a Q-guided CVAE model help to obtain high-value policy priors?
(3) Can the EBP demonstrate robustness under inaccurate value estimation?
(4) Does the algorithmic module around environment-specific design choices lead to consistently reliable performance?
For (1), we compared EBP to those that have been designed to use policy prior to accelerate online learning. For (2), we perform an analysis to demonstrate the importance of using Q-guided loss. Then, for (3), we conducted a comparative analysis of performance under varying percentages of reward noise perturbation. Lastly, for (4), we conducted experiments using the EBP algorithm with a fixed set of parameters across three challenging environments.

\subsection{Experimental Setup}
\label{sec:alg_detaild}
\textbf{Hyperparameters.}
Our proposed EBP algorithm is implemented based on TD3, and to ensure fair comparison, we adopt the hyperparameters suggested in the original paper, setting $\epsilon$=0.2 and $c$=0.5.  
In Section~\ref{sec:hpg}, EBP involves three key hyperparameters: the Q-guided coefficient $\alpha$ in the CVAE, the number of policy priors $H$, and the EPG coefficient $\mu$. To progressively emphasize Q-guided policy updates during online RL training, we additionally introduce an EPG exponential decay parameter, $\text{decay\_rate}$, to gradually attenuate the influence of EPG over time. The evolution of the EPG coefficient $\mu$ during training follows the schedule described in Eq.~\ref{eq:decay_}. For all tasks, we set $\alpha$=0.11, $H$=10, $\mu$=1.0, and $\text{decay\_rate}$=0.97. All hyperparameters for EBP, EBP(DDPG), and baselines are detailed in Table~\ref{all_hyper}. 
\begin{equation}
\label{eq:decay_}
\mu =  
\begin{cases}
\text{decay\_rate} \cdot \mu,& \text{time\_step}\% 20000 = 0 \\
     \mu, & \text{otherwise}  .
\end{cases}
\end{equation}
   
Unless otherwise stated, we maintain consistent hyperparameter configurations across all experiments. 
Regarding model architecture, our method faithfully replicates the structural framework of the TD3~\cite{fujimoto2018addressing} algorithm.
All experiments are conducted in a Linux environment equipped with a 56-core Xeon(R) 6133 CPU and 1 Nvidia RTX 3090 GPU.
The detailed experimental parameters used in the three continuous control environments can be found in Section~\ref{appendix_datatset}.

\textbf{Software.}
We use the following software versions:
\begin{itemize}
    \item Python 3.9
     \item Pytorch 2.5.0
     \item Gym 0.26.2
     \item MuJoCo 3.3.1
     \item MuJoCo-py 2.1.2.14
      \item PyBullet 3.2.7
       \item dm\_control  1.0.28

\end{itemize}

\textbf{Random seeds.}
To ensure the reproducibility of EBP, we evaluated each algorithm using ten random seeds. Furthermore, we maintained consistent seeds across all experiments, applying them to PyTorch, Numpy, Gym, and CUDA packages.

\textbf{Evaluation.}
No data or parameters were repeated for training in the evaluation, and each evaluation step consisted of 10 randomly initialized rounds. We train for 2 million time steps and evaluate the policy every 5000 time steps. 
We present the mean and standard deviation of the final 10 trials.

\textbf{Baselines.}
To evaluate the effectiveness of our method, we compare it with representative baselines, including deterministic and stochastic policy gradient algorithms (DDPG~\cite{lillicrap2015continuous}, TD3~\cite{fujimoto2018addressing}, SAC~\cite{Haarnoja2018SAC}), the blended exploitation and exploration actor-critic (BAC~\cite{ji2023seizing}), augmented state representation learning (ALH~\cite{quangaugmenting}), and nearest-neighbor–guided policy distillation (NNPG~\cite{shen2021theoretically}). For fairness, all baselines are evaluated using carefully tuned hyperparameters, detailed in Table~\ref{all_hyper} of the Appendix.

(1) DDPG~\cite{lillicrap2015continuous}: DDPG addresses the optimization challenge in continuous action spaces through the Actor-Critic framework and deterministic policy gradients, while enhancing stability via experience replay and target networks.

(2) TD3~\cite{fujimoto2018addressing}: TD3 stabilizes training by incorporating delayed policy updates and target policy smoothing regularization. Both DDPG and TD3 operate without leveraging any policy prior knowledge, relying solely on online interactions to learn an optimal policy.

(3) SAC~\cite{Haarnoja2018SAC}: SAC is a stochastic policy algorithm built upon the maximum entropy reinforcement learning framework, which achieves efficient learning in continuous action spaces by maximizing a trade-off between expected return and policy entropy.

(4) BAC~\cite{ji2023seizing}: BAC introduces a blended exploitation and exploration (BEE) operator, which jointly leverages a state–action value network ($Q$) and a state value network ($V$) to construct the target $Q$ values. This design enables the learned policy to maintain optimistic exploration while simultaneously exploiting both historically best-performing actions and the current policy during value updates.

(5) ALH~\cite{quangaugmenting}: ALH is an improvement built upon the TD3 algorithm, serving as a decision enhancement method that leverages state representations from the previous timestep to improve the perceptual capability of the Actor network and enhance the decision-making process.

(6) Nearest Neighbor Policy Guidance (NNPG): We implement the nearest-neighbor exploration algorithm proposed in NNAC~\cite{shen2021theoretically} based on the TD3 algorithm. NNPG employs the nearest neighbor action from the online replay buffer as an anchor for the current policy decision and adopts a supervised learning paradigm to guide the actor update. The overall actor loss can be formulated as
\begin{equation}
    \label{nn_exploreation_baseline1}
    \begin{aligned}
J_{\pi}(\phi)= {\mathbb{E}_{s}}\left[- {{Q_\theta }(s,{\pi _\phi }(s)) + \mu {{({\pi _\phi }(s) - {{\tilde a}})}^2}} \right].
\end{aligned}
\end{equation}

We introduce an exponential decay parameter, $\text{decay\_rate}$, as defined in Eq.~\ref{eq:decay_}, to gradually reduce the influence of the nearest neighbor actions over time, similar to the EBP algorithm.
The nearest neighbor action $\tilde a$ is queried from the replay buffer $\B_a$, which can be formulated as
\begin{equation}
     {\tilde a}  = {{\mathop{\argmin}_{a}}}(||\pi_\phi(s)-\B_{a}||_{\infty}) ,
\end{equation}
where $||\cdot||_{\infty}$ denotes the Chebyshev distance~\cite{deza2009encyclopedia}.


(7) EBP(DDPG): We implement the EBP algorithm based on the DDPG framework, utilizing a single Q-network for value estimation, denoted as EBP(DDPG). The NNPG, EBP(DDPG), and EBP algorithms incorporate supervised learning techniques to refine policy updates, aiming to enhance sample efficiency and improve policy exploitation.

(8) EBP: Built upon the TD3, the EBP algorithm additionally trains a Q-CVAE network, which provides expert prior guidance for policy improvement. Subsequently, EBP leverages a support set expert policy to guide the actor update, thereby correcting the gradient direction induced by the Q-network and enabling more stable policy updates for the RL agent.

\begin{figure*}[!htbp]
\begin{center}
\centerline{\includegraphics[width=1.0\textwidth]{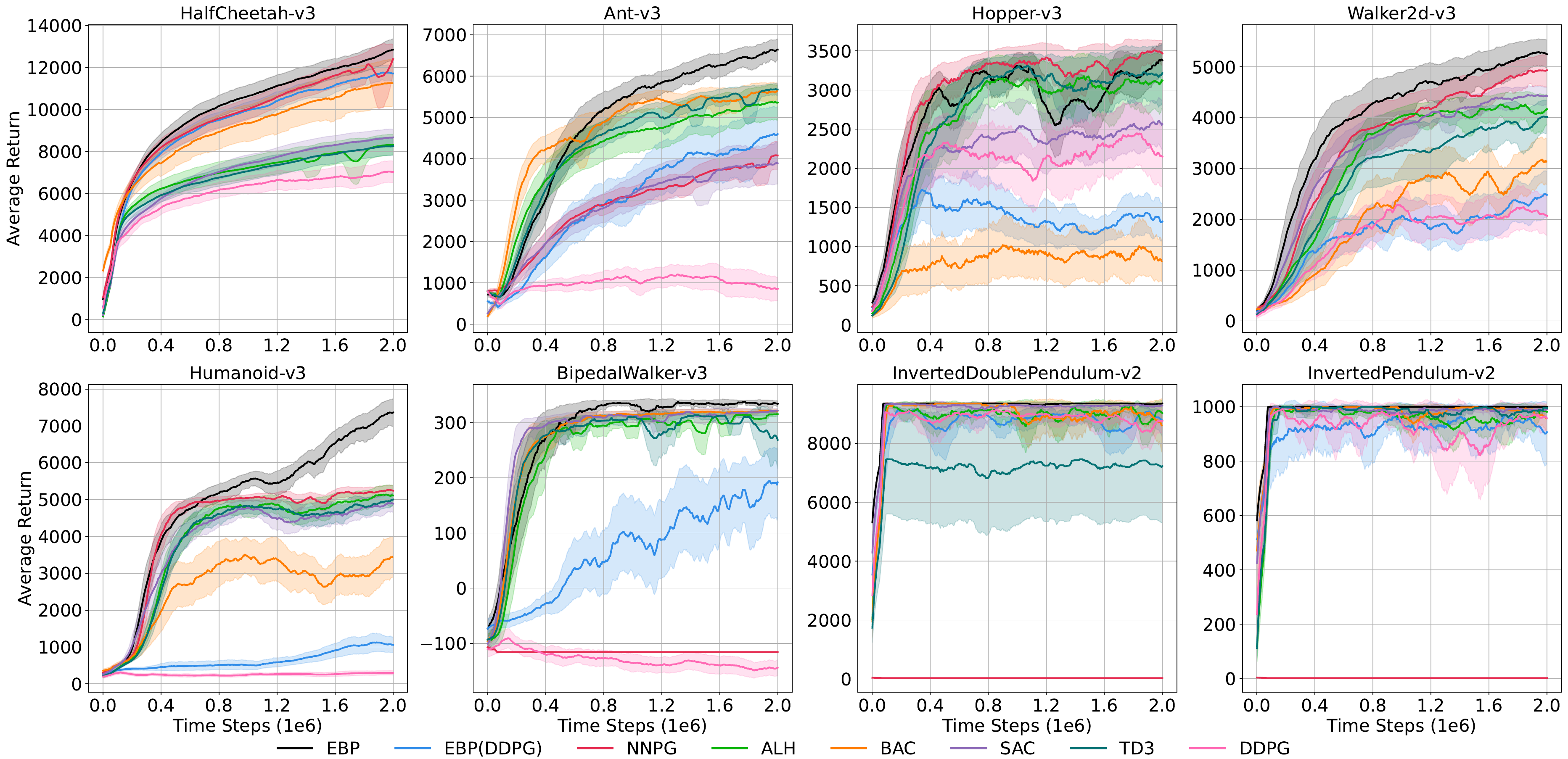}}
\caption{Learning curves on eight Gym continuous control tasks. The shaded region represents half a standard deviation of the average evaluation over 10 trials. Curves are smoothed uniformly for visual clarity.}
\label{fig:Over_results}
\end{center}
\vskip -0.2in
\end{figure*}

\subsection{Experiments on Gym Environment}

\textbf{Can EBP leverage behavior policy to accelerate online learning?}
We compare our proposed EBP algorithm to these baselines, and these comparisons are performed on eight Gym continuous control tasks. 
The learning curves are presented in Fig.~\ref{fig:Over_results}, while both final and intermediate results are reported in Table~\ref{tab:mojoco_res}.

As shown in Fig.~\ref{fig:Over_results}, the proposed EBP algorithm consistently outperforms the vanilla TD3 across eight different tasks, demonstrating significant advancements and stability compared to other policy-prior guidance methods (NNPG, ALH).
Specifically, on the benchmark tasks BipedalWalker,  InvertedDouble (denoted as InvertedDoublePendulum), and InvertedPendulum, the NNPG fails to learn effective control policies.
We attribute this to the lack of diverse expert policy anchors serving as optimization references in the early training phase of the NNPG, which results in overly conservative policy anchors being provided for relatively easy tasks (with action dimensions of 4, 1, and 1, respectively).
We argue that policies in easy tasks converge more rapidly, thereby making high-value anchors especially critical for effective guidance in such tasks.
The ALH method generates the current action based on a combination of the previous timestep's observation representation and the current state representation, which can result in similar consecutive actions and consequently lead to suboptimal performance.

In contrast, the EBP algorithm significantly improves performance by generating diverse, high-value policy priors to guide the policy update process.
Additionally, we observe that the BAC algorithm exhibits noticeable performance degradation on the Hopper, Walker2d, and Humanoid tasks compared to SAC. We conjecture that excessive reliance on historical policies may bias action selection toward exploitation, thereby impairing effective exploration in environments with complex and highly dynamic behaviors.

 
 





\begin{table*}[ht]
\centering
\setlength\tabcolsep{5pt}
\caption{Average performance on the Gym benchmark at 200k, 1M, and 2M timesteps over 10 random seeds. $\pm$ corresponds to a standard deviation over trials. The highest performance is bolded.}
\begin{tabular}{lccccccc|cc}
\toprule
Tasks      & Time Steps&DDPG&TD3&SAC&BAC&ALH&NNPG&EBP(DDPG)&EBP   \\  \midrule
\multirow{3}{*}{HalfCheetah-v3}&200K& 4193$\pm$335&4977  $\pm$243&4516$\pm$210&6039$\pm$425&5208$\pm$164&\bf 6204$\pm$276&5832$\pm$293&6177$\pm$220\\
&1M& 6387$\pm$402  &7030$\pm$440&7397$\pm$398&9346$\pm$849&7180$\pm$524&9970$\pm$502&9941$\pm$394&\bf 10597$\pm$540\\
&2M& 7030$\pm$512  &8259$\pm$430&8672$\pm$399&11255$\pm$1134&8339$\pm$484&12429$\pm$695&11739$\pm$382&\bf 
12877$\pm$516\\
 \midrule
 \multirow{3}{*}{Ant-v3}&200K&794$\pm$76 &1308$\pm$191& 1036$\pm$172&\bf 2505$\pm$473&1782$\pm$365&1054$\pm$125&700$\pm$125&1070$\pm$223 \\
&1M& 1142$\pm$239&4820$\pm$269&3074$\pm$598&5327$\pm$189&4640$\pm$495&3088$\pm$228&3119$\pm$588&\bf 5474$\pm$418 \\
&2M& 848$\pm$286 &5697$\pm$134&3906$\pm$508&5631$\pm$205&5391$\pm$427&4083$\pm$347&4579$\pm$682&\bf 6672$\pm$264 \\
 \midrule
\multirow{3}{*}{Hopper-v3}&200K& 1216$\pm$312&653$\pm$169&1439$\pm$354&568$\pm$257&1010$\pm$369&1593$\pm$167&1191$\pm$317&\bf 2303$\pm$443 \\
&1M& 2076$\pm$424 &3273$\pm$184&2463$\pm$392&1018$\pm$404&3001$\pm$392&3312$\pm$229&1382$\pm$287&\bf 3448$\pm$131 \\
&2M& 2150$\pm$383 &3216$\pm$372&2552$\pm$320&806$\pm$274&3130$\pm$367&\bf 3467$\pm$168&1343$\pm$254&3279$\pm$256 \\
 \midrule
\multirow{3}{*}{Walker2d-v3}&200K& 409$\pm$113&653$\pm$154 &982$\pm$279&364$\pm$57&675$\pm$242&686$\pm$180 &491$\pm$95&\bf 1447$\pm$390\\
&1M& 2243$\pm$377&3318$\pm$570 &3813$\pm$293&2297$\pm$549&4005$\pm$189&3834$\pm$508 &2074$\pm$350&\bf 4429$\pm$422\\
&2M& 2045$\pm$389& 4008$\pm$322 &4419$\pm$320&3102$\pm$468&4159$\pm$303&4978$\pm$291 &2467$\pm$487&\bf 5247$\pm$283\\
 \midrule

\multirow{3}{*}{Humanoid-v3}&200K& 281$\pm$36 &577$\pm$33&\bf 753$\pm$126&627$\pm$60&611$\pm$41&583$\pm$20&404$\pm$35& 711$\pm$46 \\
&1M& 238$\pm$39 &4844$\pm$187&4775$\pm$229&3445$\pm$576&4842$\pm$248&5053$\pm$53&520$\pm$127&\bf 5446$\pm$228 \\
&2M& 297$\pm$60 &5003$\pm$169&4892$\pm$230&3444$\pm$571&5148$\pm$273&5292$\pm$88&1057$\pm$205&\bf 7359$\pm$380\\
 \midrule
\multirow{3}{*}{BipedalWalker-v3}&200K& -96$\pm$17 & 118$\pm$63&\bf 231$\pm$50&135$\pm$66&54$\pm$69&-116$\pm$0&-59$\pm$13&93$\pm$74 \\
&1M& -137$\pm$16  &310$\pm$5&313$\pm$9&315$\pm$7&288$\pm$27&-116$\pm$0&101$\pm$71&\bf333$\pm$7 \\
&2M& -146$\pm$14  &265$\pm$57&322$\pm$4&321$\pm$4&316$\pm$4&-116$\pm$0&190$\pm$65&\bf 334$\pm$6\\
 \midrule
\multirow{3}{*}{InvertedDouble-v2}&200K& 8843$\pm$445&\ 7368$\pm$1877&9341$\pm$2&9359$\pm$1&9233$\pm$123&28$\pm$0&9163$\pm$234&\bf 9356$\pm$2\\
&1M& 9016$\pm$253&\ 7037$\pm$1930&9321$\pm$30&9207$\pm$212&9064$\pm$373&28$\pm$0&8898$\pm$306&\bf 9360$\pm$0\\
&2M& 8766$\pm$385&\ 7293$\pm$1876&9293$\pm$54&8628$\pm$877&9088$\pm$359&28$\pm$0&8725$\pm$605&\bf 9358$\pm$3\\
 \midrule
\multirow{3}{*}{InvertedPendulum-v2}&200K& 991$\pm$11&\bf 1000$\pm$0&\bf 1000$\pm$0&999$\pm$0&988$\pm$18&2$\pm$0 &900$\pm$107&\bf 1000$\pm$0  \\
&1M& 972$\pm$42& 999$\pm$1&959$\pm$52&999$\pm$0&976$\pm$31&2$\pm$0 &954$\pm$64&\bf 1000$\pm$0  \\
&2M& 954$\pm$60&995$\pm$6&991$\pm$11&954$\pm$55&985$\pm$19&2$\pm$0 &915$\pm$105&\bf 1000$\pm$0  \\
\bottomrule
\end{tabular}
\label{tab:mojoco_res}
\end{table*}
As shown in Table~\ref{tab:mojoco_res}, compared with the vanilla TD3 algorithm, EBP has an average performance gain of 51.2\%, 19.5\%, and 26.0\% at 200K, 1M, and 2M timesteps, respectively.
Meanwhile, EBP(DDPG) outperforms the original DDPG algorithm by 5.6\%, 16.2\%, and 63.9\% at 200K, 1M, and 2M timesteps, respectively.
Although ALH and NNPG incorporate historical state representations and adopt more computationally intensive architectures, separately, EBP significantly outperforms these baselines in both early-stage performance (200K timesteps) and final performance (2M timesteps). Notably, EBP often surpasses the performance of TD3 at 2M timesteps as early as 1M timesteps, highlighting a substantial performance gain.

\subsection{Experiments on PyBullet Environment}

\begin{figure*}[!htbp]
\begin{center}
\centerline{\includegraphics[width=1.0\textwidth]{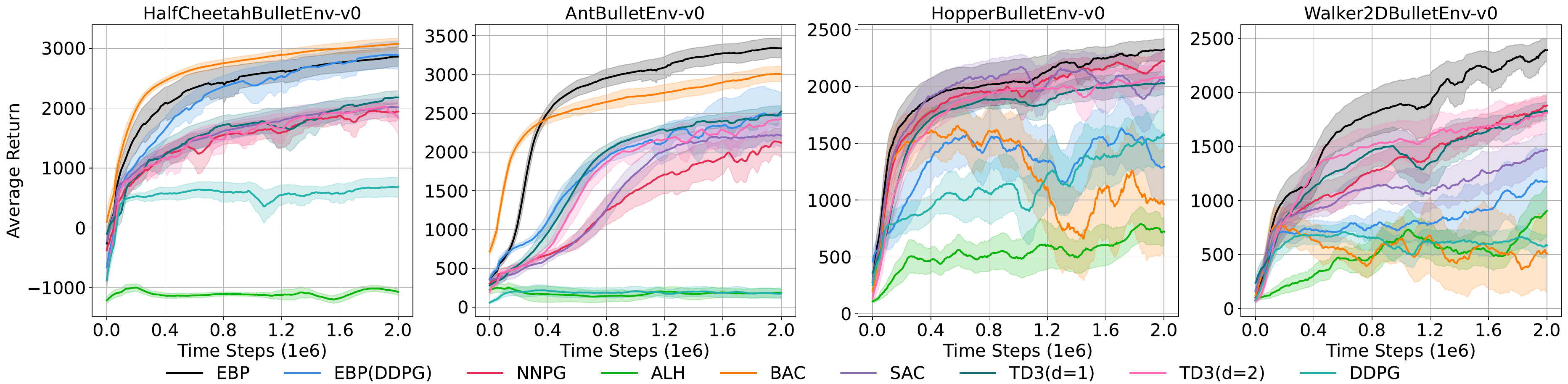}}
\caption{Learning curves on four PyBullet continuous control tasks. The shaded region represents half a standard deviation of the average evaluation over 10 trials. Curves are smoothed uniformly for visual clarity.}
\label{fig:Over_results_pybullet}
\end{center}
\vskip -0.2in
\end{figure*}

\textbf{Experiments on Four PyBullet Tasks.}
We evaluated the effectiveness of the EBP algorithm on the state-based PyBullet benchmark suite, with results presented in Fig.~\ref{fig:Over_results_pybullet}. 

The experimental results demonstrate that: (1) EBP, implemented on top of TD3, consistently outperforms the vanilla TD3 across all four tasks; and (2) EBP further achieves superior performance compared to NNPG. 
We additionally observe that the BAC algorithm performs favorably on the HalfCheetahBulletEnv and AntBulletEnv tasks, but exhibits degraded performance on HopperBulletEnv and Walker2DBulletEnv, indicating notable performance instability across environments. We attribute this behavior to BAC’s excessive reliance on early-stage experiences, which may bias policy updates toward exploitation and limit optimistic exploration, particularly in tasks with sensitive dynamics or narrow stability regions. This trend is consistent with BAC’s performance in the Hopper and Humanoid tasks in the Gym benchmark.


At the same time, we constructed EBP(DDPG) based on the DDPG, using the same set of hyperparameters as in EBP, as summarized in Table~\ref{all_hyper}. EBP(DDPG) demonstrates notably superior performance compared to vanilla DDPG across multiple aspects: in all four PyBullet environments, EBP(DDPG) significantly outperforms vanilla DDPG.
In the HalfCheetahBulletEnv and AntBulletEnv tasks, the proposed EBP(DDPG) even outperforms TD3, further validating the positive effect of expert policy priors in facilitating policy updates. In the Walker2DBulletEnv task, the vanilla DDPG algorithm tends to prematurely converge to a local optimum, while the expert-prior-guided EBP(DDPG) enables further policy improvement.
These significant improvements highlight that EBP can effectively enhance the practical performance of online RL algorithms.

\textbf{Effectiveness of Increasing Actor Update Frequency.}
To validate the effectiveness of the synchronized policy update mechanism, we conducted a systematic comparison of TD3 under different policy update configurations. Our proposed EBP algorithm was evaluated under both synchronized updates ($d$=1) and delayed updates ($d$=2) during the training phase, as illustrated in Algorithm~\ref{alg:policy_update}. 

\begin{center}  
 
\begin{minipage}{.35\textwidth} 
\begin{algorithm}[H] 
   \caption{Policy Update Mechanism}
   \label{alg:policy_update}
\begin{algorithmic}
   \STATE Initialize timestep=0, $d$=1 or 2
   \WHILE {True}
   \STATE timestep=timestep+1
   \STATE Train critics
   \IF {timestep\%\textit{d}==0} 
   \STATE Train actor
   \STATE Update weights
   \ENDIF
   \ENDWHILE
\end{algorithmic}
\end{algorithm}
\end{minipage}
\end{center}

As shown in Fig.~\ref{fig:Over_results_pybullet}, EBP consistently outperforms the vanilla TD3 across multiple tasks, achieving significantly faster convergence. Notably, on the HalfCheetahBulletEnv, AntBulletEnv, and Walker2DBulletEnv tasks, TD3 with synchronized updates ($d$=1) outperforms the delayed update setting ($d$=2), suggesting that synchronized updates can enhance policy exploitation. Furthermore, EBP surpasses both TD3($d$=1) and TD3($d$=2) in all four tasks, demonstrating its superior capability in efficient policy exploitation.

 \begin{figure*}[!htbp]
\begin{center}
\centerline{\includegraphics[width=1.0\textwidth]{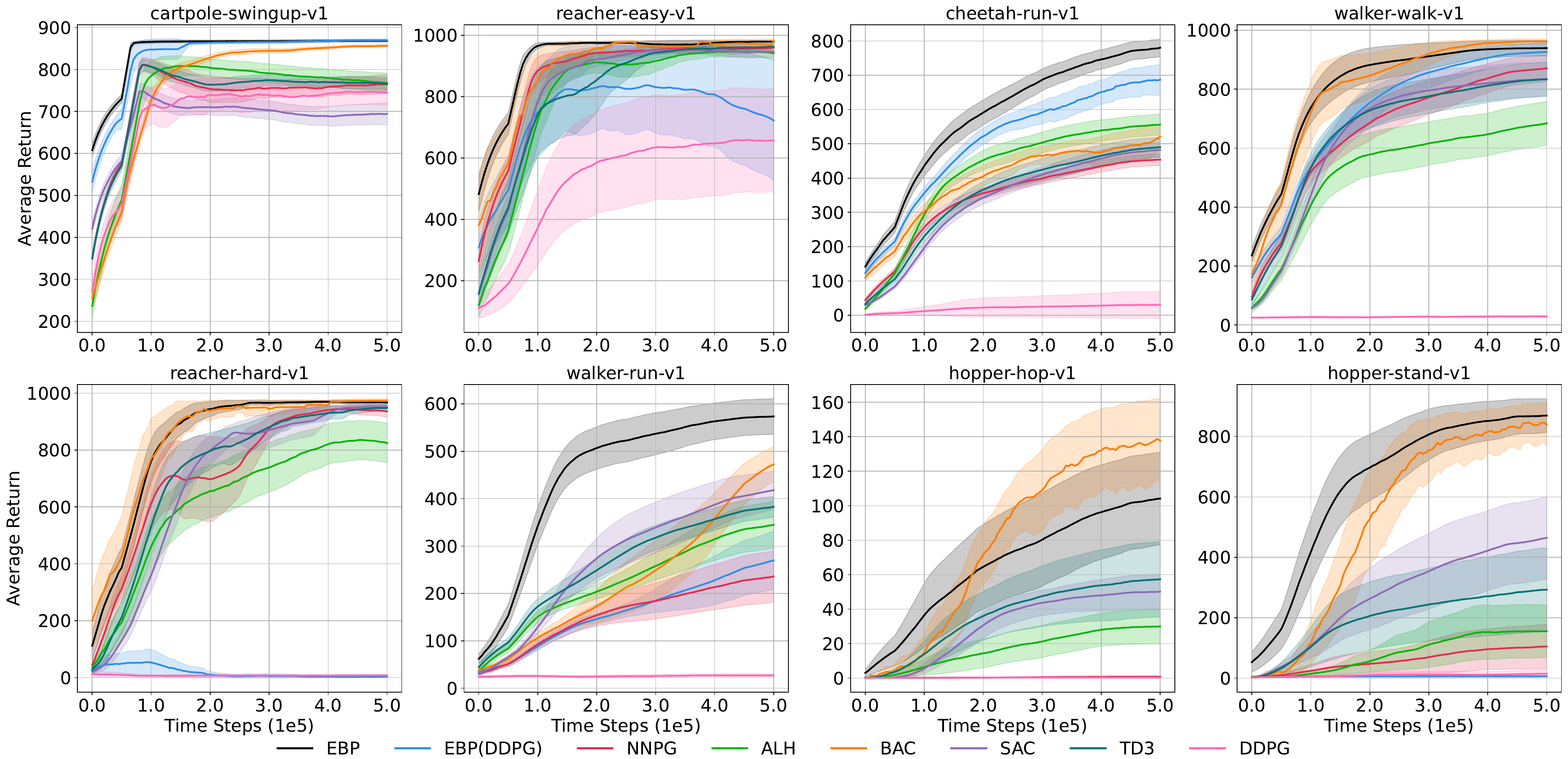}}
\caption{Learning curves on four DMControl continuous control tasks. The shaded region represents half a standard deviation of the average evaluation over 10 trials. Curves are smoothed uniformly for visual clarity.}
\label{fig:Over_results_dmc}
\end{center}
\vskip -0.2in
\end{figure*}
\subsection{Experiments on DMControl Suite}
We present the performance curves of EBP  on a total of eight DMControl environments in Fig.~\ref{fig:Over_results_dmc}. We run 10 seeds in each environment.
Experimental results demonstrate that the proposed EBP algorithm significantly accelerates policy convergence. Across all eight evaluated tasks, both EBP and EBP(DDPG) consistently outperform the vanilla TD3 and DDPG algorithms, respectively, further demonstrating the effectiveness and generality of our approach.

In particular, on the walker-walk, cheetah-run, reacher-hard, walker-run, hopper-hop, and hopper-stand tasks, DDPG, which relies on a single Q-network for policy updates, fails to learn effective policies, resulting in persistently low returns throughout the training process. This is primarily due to the inherent overestimation bias and instability of the single-critic architecture in high-dimensional continuous control tasks. In contrast, EBP(DDPG) introduces an expert prior-guided mechanism, where a generative expert prior is leveraged to assist policy updates, providing more stable and informative gradients for policy learning and effectively alleviating the aforementioned issues. As a result, EBP(DDPG) achieves significantly higher returns and faster convergence on both tasks, demonstrating the effectiveness of the proposed method in improving the performance of the vanilla DDPG algorithm.

\begin{figure}
    \centering
    \includegraphics[width=0.5\textwidth]{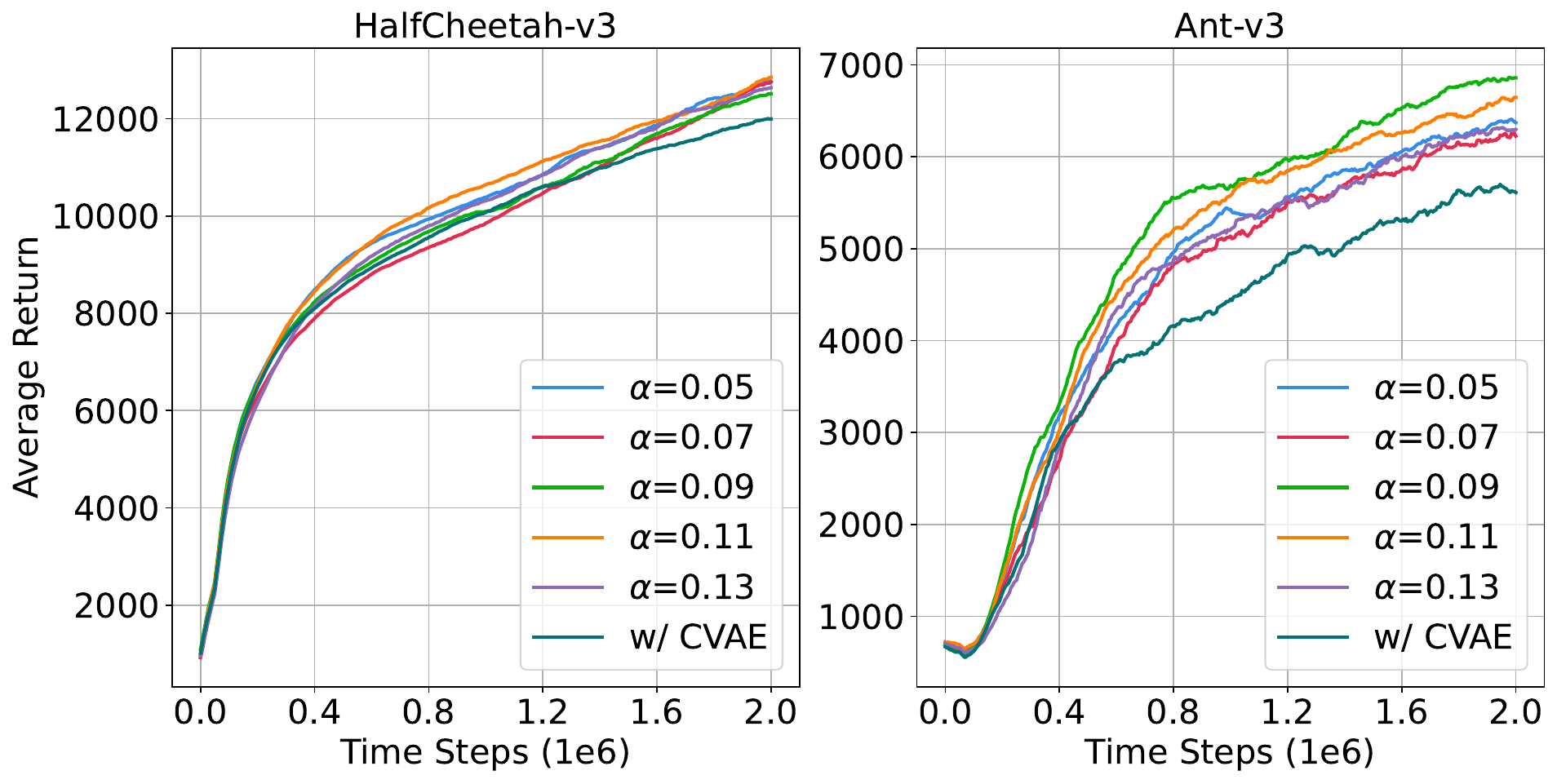}
\caption{Comparison of the average returns obtained by the Q-CVAE model with different Q-guided coefficients $\alpha$ on the HalfCheetah and Ant tasks. Curves are smoothed uniformly for visual clarity.}
    \label{ab_qcave}
\end{figure}

\subsection{EBP Analysis and Ablation Study}
\label{sec_abla}
\textbf{Can a Q-guided CVAE model help to obtain high-value policy priors?}
To analyze the effect of different $\alpha$ values of the Q-CVAE on the EBP algorithm, we conducted a series of experiments. We evaluated them on the HalfCheetah and Ant tasks using $\alpha$ values of $(0.05, 0.09, 0.11, 0.13)$. Subsequently, we compared the results of EBP with those of EBP without the Q-guided loss in the CVAE, denoted as EBP(w/ CVAE), and the results are shown in Fig.~\ref{ab_qcave}.

From the experimental results in Fig.~\ref{ab_qcave}, we draw two key observations. First, both EBP(w/ Q-CVAE) and EBP(w/ CVAE) outperform the vanilla TD3, indicating that leveraging prior policies to guide the update process can improve performance. Second, EBP(w/ Q-CVAE) with varying values of the weighting parameter $\alpha$ consistently enhances performance, demonstrating the effectiveness of the Q-guided loss term $\mathcal{H}_{\text{Q}}(\omega)$ in producing higher-quality actions. This highlights the importance of providing explicit policy guidance during the training phase. Moreover, compared with vanilla TD3, the EBP algorithm shows robustness to the choice of $\alpha$, with values ranging from 0.05 to 0.13, all leading to significant performance gains.

\begin{figure}[!htbp]
    \centering
    \includegraphics[width=0.5\textwidth]{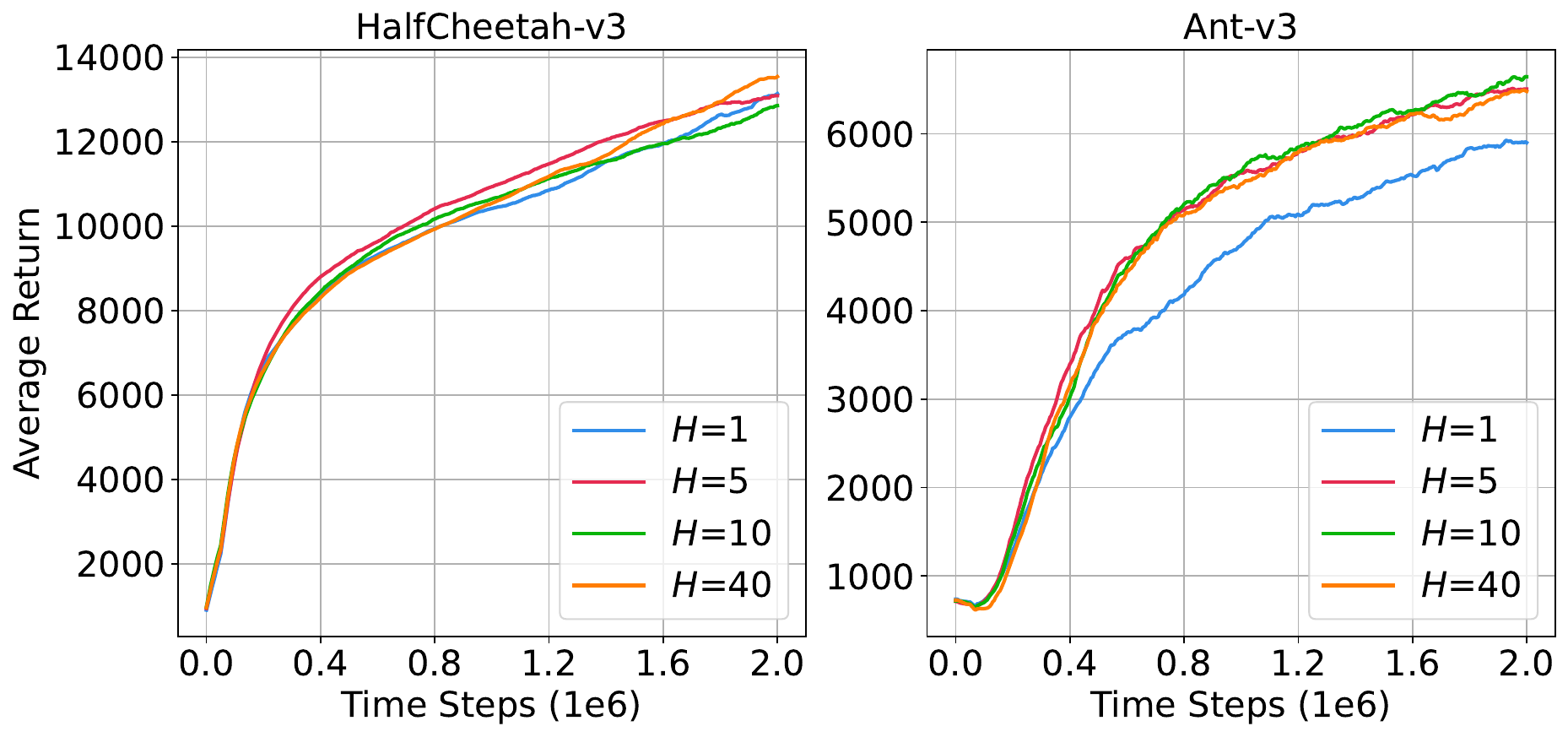}
    \caption{Ablation experiments of our algorithm with different numbers of policy priors $H$ on HalfCheetah and Ant tasks. Curves are smoothed uniformly for visual clarity.}
    \label{fig:policy_prior_guidance_H}
\end{figure}
\textbf{Effectiveness of Expert Policy Guidance.}
We conducted a systematic ablation analysis on Gym continuous control environments to evaluate two configuration variants, aiming to quantify the contribution of EPG in the EBP algorithm. Therefore, we performed ablation studies on two critical parameters: the number of policy priors $H$ and the expert policy constraint strength $\mu$. 

First, we evaluated the impact of varying $H$ on the HalfCheetah and Ant task, using $H$=(1, 5, 10, 40) with 10 random seeds per setting.
In these settings, both Q-values and generative policy priors are leveraged to guide the policy update process.
The results demonstrate that expert prior effectively enhances performance, as illustrated in Fig.~\ref{fig:policy_prior_guidance_H}.
We observe that EBP significantly improves the performance of the vanilla TD3 algorithm across varying numbers of policy priors $H$. This indicates that incorporating expert policy priors as an update anchor can effectively mitigate the issue of suboptimal exploration induced by Q-guidance. However, as $H$ increases, the likelihood of sampling previously seen low-quality actions also increases, which can introduce instability during training. Based on this empirical observation, we set $H = 10$ in all experiments to balance performance gains and training stability.

\begin{figure}[!htbp]
    \centering
    \includegraphics[width=0.5\textwidth]{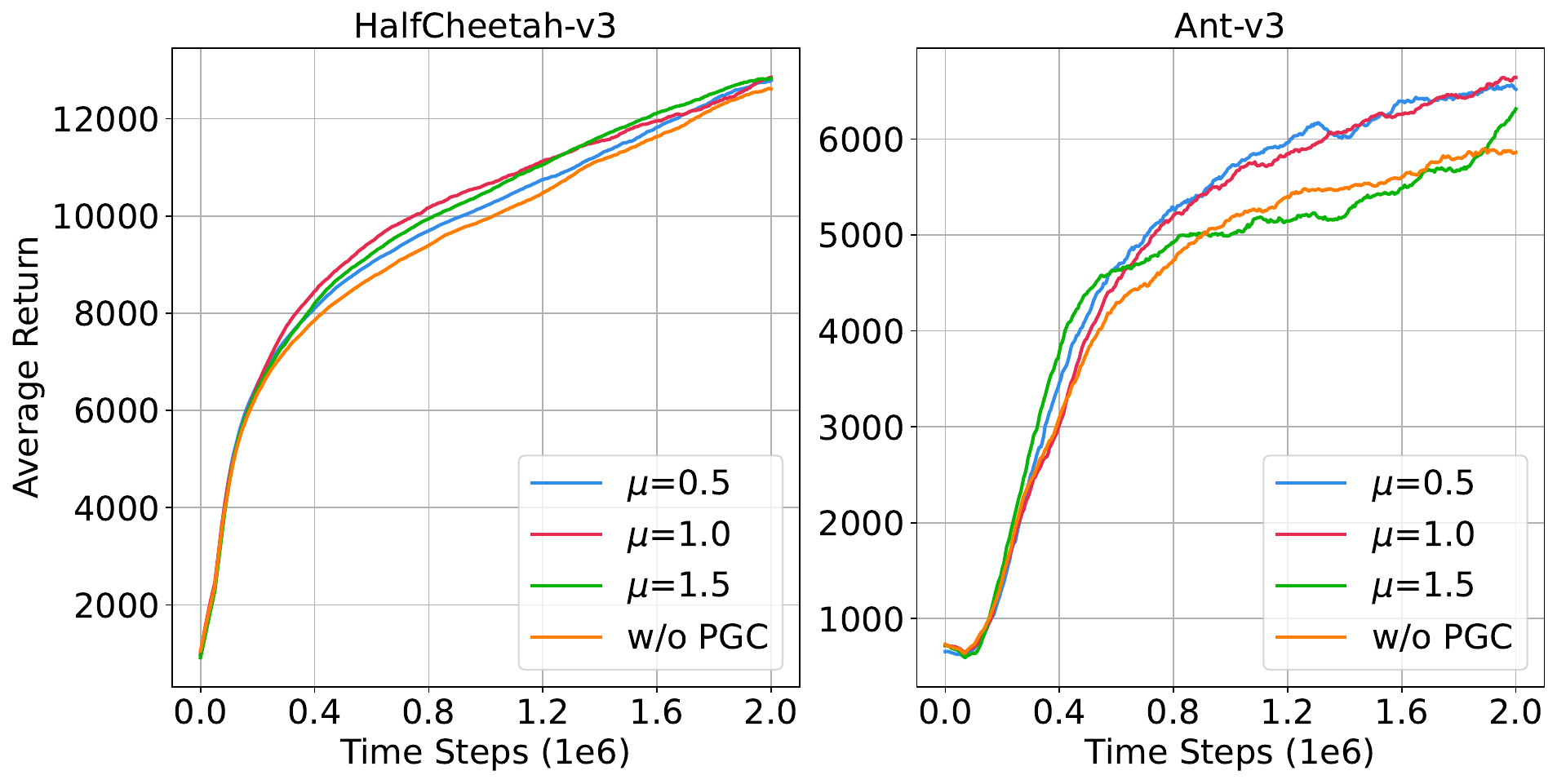}
    \caption{Ablation experiments of our algorithm with different PGC coefficients $\mu$ on HalfCheetah and Ant tasks. Curves are smoothed uniformly for visual clarity.}
    \label{fig:policy_prior_guidance_mu}
\end{figure}

Furthermore, we conducted experiments during the training phase with three different values of $\mu$=(0.5, 1.0, 1.5). Each model is trained for 2 million time steps on the HalfCheetah and Ant task, using 10 random seeds. The results are presented in Fig.~\ref{fig:policy_prior_guidance_mu}.
Notably, when $\mu = 0.5$ or $\mu = 1.0$, the proposed method achieves consistently strong performance on both HalfCheetah and Ant. In contrast, setting $\mu = 1.5$ leads to noticeably slower performance improvements in the later stages of training. 
This suggests that an overly large $\mu$ introduces excessive policy conservatism, thereby impeding the efficiency of Q-guided policy updates.
These results indicate that the EPG module improves performance by incorporating an auxiliary expert policy optimization objective.

\textbf{Effectiveness of Policy Gradient Correction.}
To evaluate the effectiveness of the proposed PGC method, we conduct comparative analyses on the HalfCheetah and Ant tasks by comparing the EBP algorithm without PGC (denoted as EBP (w/o PGC)) against EBP variants with different values of the parameter $\mu$. The results are shown in Fig.~\ref{fig:policy_prior_guidance_mu}.

As illustrated in Fig.~\ref{fig:policy_prior_guidance_mu}, on the HalfCheetah and Ant task, EBP (w/o PGC) consistently performs significantly worse than all EBP variants incorporating PGC, including EBP ($\mu=0.5$), EBP ($\mu=1.0$), and EBP ($\mu=1.5$). Among these variants, EBP ($\mu=1.0$) achieves the best overall performance.
We attribute this improvement to the role of the PGC module within the combined EPG and Q-guidance framework. Specifically, PGC introduces a balanced dual-objective optimization mechanism that effectively alleviates gradient conflicts during training, thereby resulting in stable and consistent performance gains.

\begin{figure}[!htbp]
    \centering
    \includegraphics[width=0.5\textwidth]{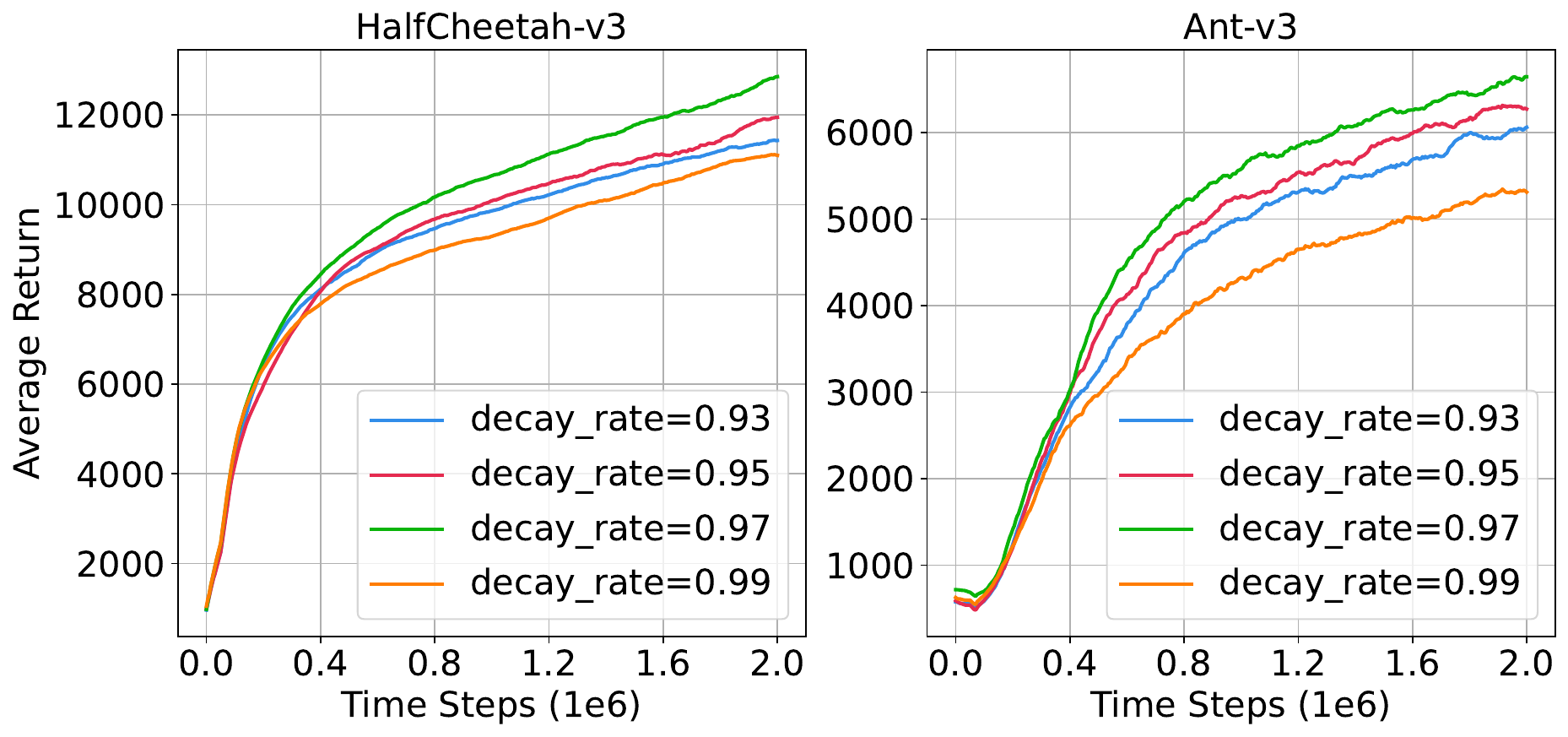}
    \caption{Ablation experiments of our algorithm with PGC coefficients $\mu = 1.0$, using different exponential decay parameters $\text{decay\_rate}$ on the HalfCheetah and Ant tasks. The curves are smoothed uniformly for visual clarity.}
    \label{fig:policy_decay_rate}
\end{figure}

Finally, we evaluate the sensitivity of the EBP algorithm to the parameter $\text{decay\_rate}$. We conduct experiments with different values of $\text{decay\_rate} = (0.93, 0.95, 0.97, 0.99)$, and the results are reported in Fig.~\ref{fig:policy_decay_rate}.

As shown in Fig.~\ref{fig:policy_decay_rate}, EBP achieves strong performance when $\text{decay\_rate} = (0.93, 0.95, 0.97)$, whereas a larger value, such as $\text{decay\_rate} = 0.99$, leads to noticeable performance degradation. 
We attribute this behavior to the role of $\text{decay\_rate}$ in controlling the strength of expert prior guidance. During the later stages of training, a weaker prior influence is beneficial for achieving higher performance, as an overly strong prior may restrict exploration and consequently hinder the final learning outcome.

\begin{figure}[!htbp]
    \centering
    \includegraphics[width=0.5\textwidth]{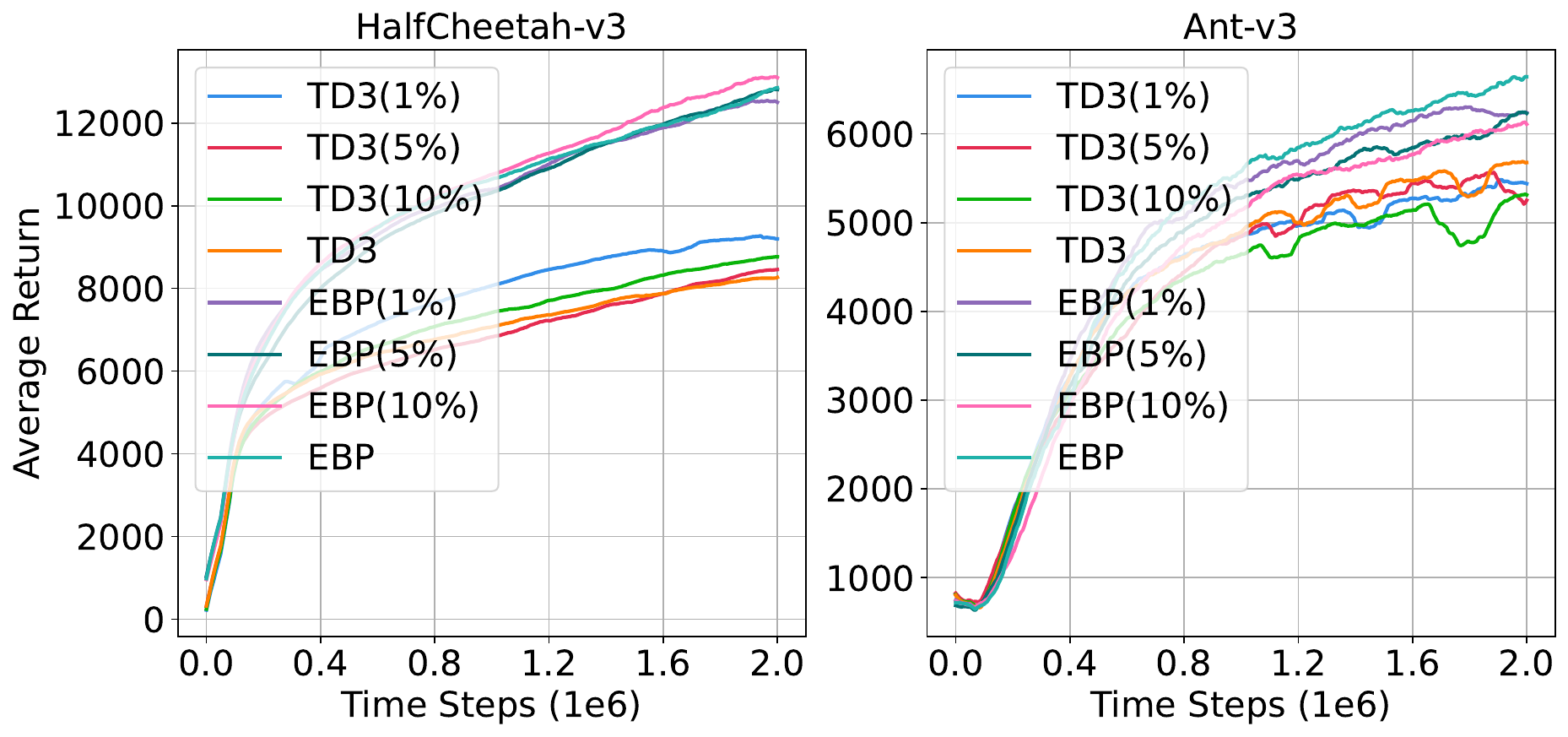}
    \caption{Average return over 10 random seeds under varying percentages of reward noise perturbation. Curves are smoothed uniformly for visual clarity.}
    \label{ab_noise_rl}
\end{figure}

\textbf{Can the EBP algorithm exhibit robustness against noise perturbations?}
In online RL, reward noise can exacerbate overestimation bias, potentially leading to training instability or even failure. To evaluate the robustness of the EBP algorithm under such perturbations, we uniformly injected reward noise throughout the training process, defined as $r^* = r + \mathcal{N}(0, \epsilon_r^2)$, where $\epsilon_r$ = 0.01.
The resulting tuples $(s, a, r^*, s')$ were stored in the replay buffer. We assess performance on the HalfCheetah and Ant task over 2M timesteps under three different noise injection rates (1\%, 5\%, 10\%), reporting the average performance over 10 random seeds. For example, a 1\% noise rate means Gaussian noise is added every 100 timesteps. The average scores are presented in Fig.~\ref{ab_noise_rl}.
 \begin{figure*}[!ht]
\begin{center}
\centerline{\includegraphics[width=1.0\textwidth]{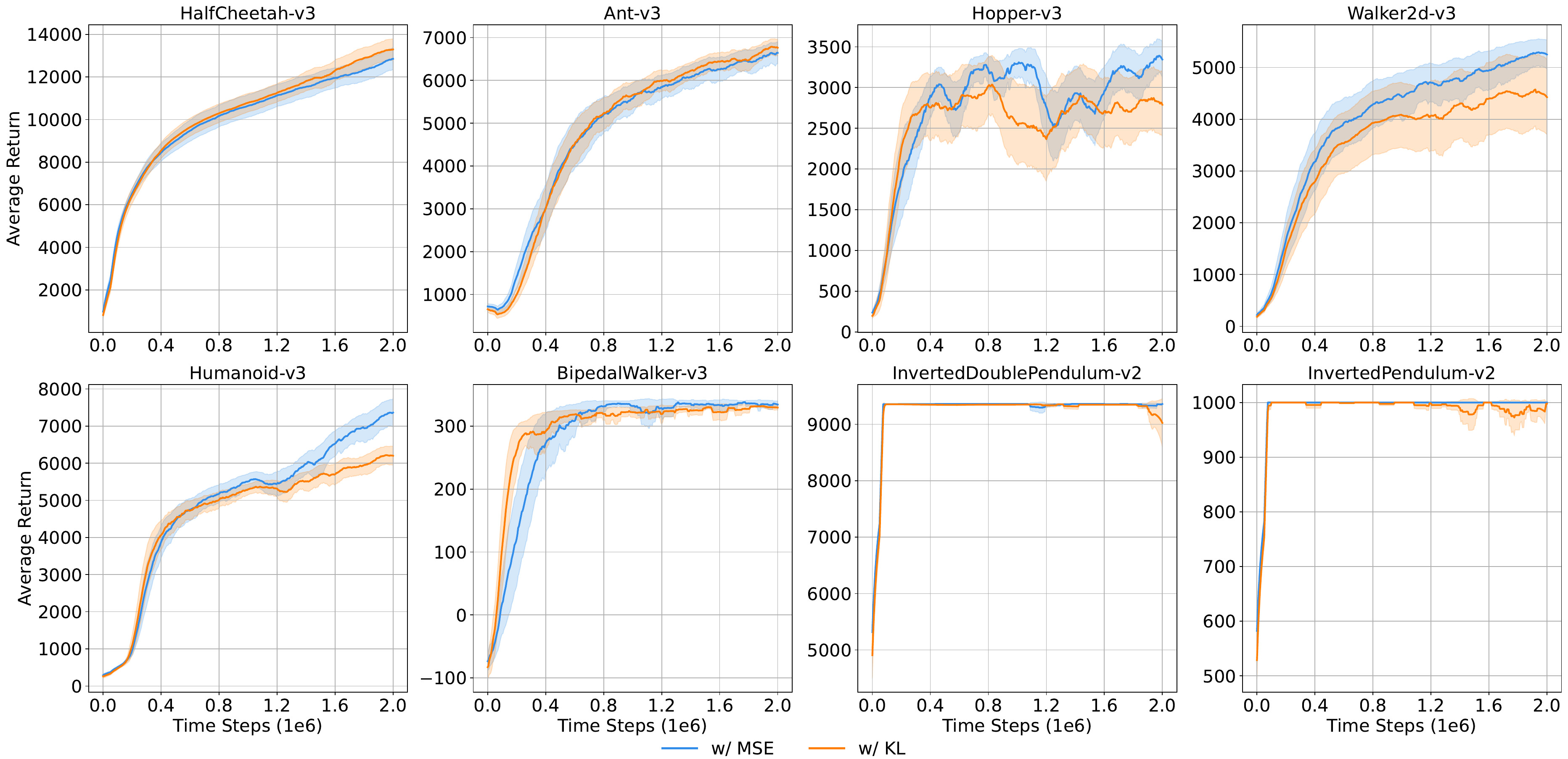}}
\caption{Comparison of learning curves obtained by using MSE and KL divergence as optimization objectives across eight Gym continuous control tasks. The shaded region represents half a standard deviation of the average evaluation over 10 trials. Curves are smoothed uniformly for visual clarity.}
\label{fig:Over_results_mujoco_KL}
\end{center}
\end{figure*}

Experimental results demonstrate a significant performance divergence between EBP and TD3 under Gaussian reward noise conditions. 
In the Ant task, both EBP and TD3 experience performance degradation when exposed to reward noise. Under 10\% noise, EBP and TD3 show a performance drop of 6\% and 8\%, respectively, compared to their noise-free counterparts (vanilla EBP and vanilla TD3). This indicates that EBP’s use of synchronized policy updates does not compromise the reward noise robustness exhibited by TD3.
In contrast, TD3 surprisingly achieves improved performance under 1\%, 5\%, and 10\% noise on the HalfCheetah task. 
We consider this phenomenon to be the smoothing effect introduced in TD3’s Q-value estimation process under noisy rewards, which effectively mitigates overestimation bias in the critic~\cite{fujimoto2018addressing}. Similar value smoothing effects have been theoretically analyzed and empirically validated to enhance training stability and reduce function approximation errors in reinforcement learning~\cite{corazza2022reinforcement,duan2022distributional}.
In summary, EBP demonstrates superior robustness in noisy environments, significantly improves sample efficiency in online RL, and accelerates policy convergence. 
This highlights the substantial potential of EBP in complex, real-world applications.

We attribute this phenomenon to the smoothing effect introduced in TD3’s Q-value estimation process under noisy rewards, which effectively mitigates overestimation bias in the critic~\cite{fujimoto2018addressing}. Similar value smoothing effects have been theoretically analyzed and empirically validated to enhance training stability and reduce function approximation errors in reinforcement learning~\cite{pan2020softmax,duan2022distributional}.

\textbf{Compare More Distance Indicators to Guide Policy Updates.}
To constrain the deviation between the agent's current policy and the expert prior, we adopt a knowledge distillation approach~\cite{sun2024logit} and conduct a comparative analysis of the EBP algorithm using two common distance metrics: Mean squared error and Kullback–Leibler divergence~\cite{hinton2015distilling}, denoted as EBP(w/ MSE) and EBP(w/ KL), respectively. The experimental results are shown in Fig.~\ref{fig:Over_results_mujoco_KL}.

It can be observed that using MSE loss as the optimization objective for EPG provides a clear advantage within the EBP algorithm, particularly on the Humanoid task. Therefore, our proposed EBP algorithm adopts MSE as the loss function for the EPG module throughout all experiments.

\begin{figure}[!htbp]
    \centering
    \includegraphics[width=0.5\textwidth]{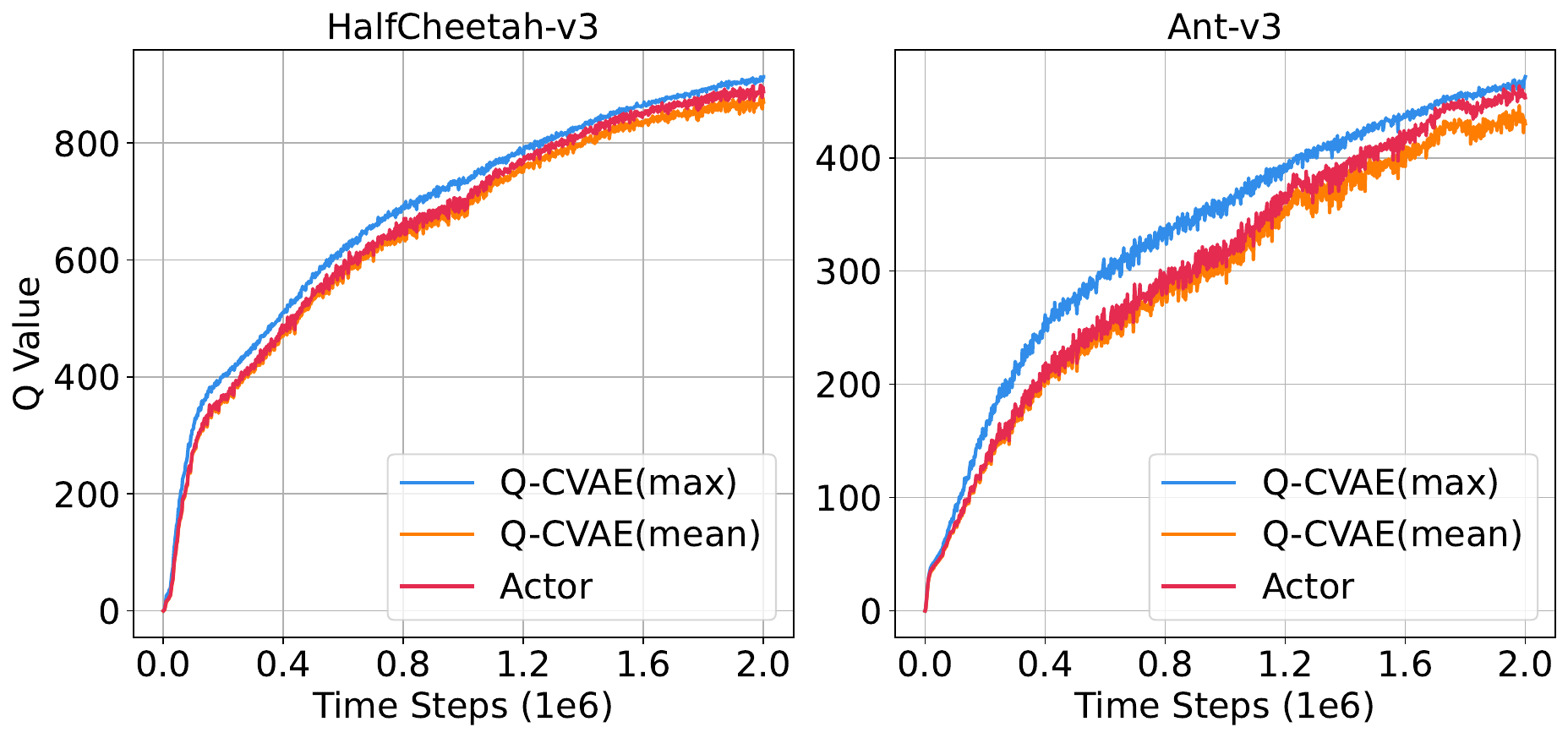}
    \caption{Q-value evolution curves for actions sampled from the Q-CVAE and the actor network on the HalfCheetah and Ant tasks. The curves are smoothed uniformly for visual clarity.}
    \label{fig:results_process}
\end{figure}
\textbf{Value Visualization of the Behavioral Policy.}
To evaluate whether the behavioral policy network is capable of generating high-value policies, we employ the EBP algorithm to assess the Q-value estimates of actions sampled from both the policy prior network $G_\omega$ and the actor network $\pi_\phi$. Specifically, we collect the maximum and mean Q-values over 10 actions sampled from the Q-CVAE, and visualize these statistics together with the Q-value trajectory of actions produced by the actor network in Fig.~\ref{fig:results_process}.

As shown in Fig.~\ref{fig:results_process}, the action-value distributions induced by the policy prior network $G_\omega$ and the actor network exhibit similar overall trends. Notably, the maximum Q-values of actions generated by the policy prior network are consistently higher than those produced by the actor network, while the mean Q-values of actions sampled from the policy prior network tend to be lower than those of the actor network in the later stages of training. This observation indicates that, throughout training, the CVAE-based behavioral policy can generate high-value trajectories, thereby providing the agent with a more informative and directive policy prior. However, in the later training phase, the gap between the Q-values of actions sampled from the Q-CVAE and those from the actor network becomes relatively small, suggesting that the guidance provided by the Q-CVAE becomes limited. Consequently, incorporating a weight decay mechanism in the EBP algorithm is necessary to appropriately attenuate the influence of the policy prior during late-stage training.

 \begin{figure*}[!tbp]
\centering
 \subfloat[7pt][Online Gym Evaluation (Full results in Fig.~\ref{fig:Over_results}).]{
		\includegraphics[height=0.135\textheight]{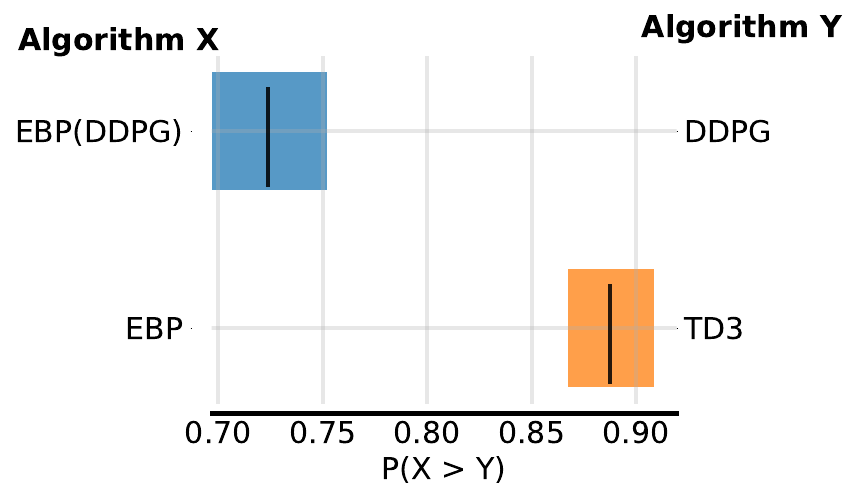}} 
\hfill
  \subfloat[8pt][Online PyBullet Evaluation (Full results in Fig.~\ref{fig:Over_results_pybullet}).]{
		\includegraphics[height=0.135\textheight]{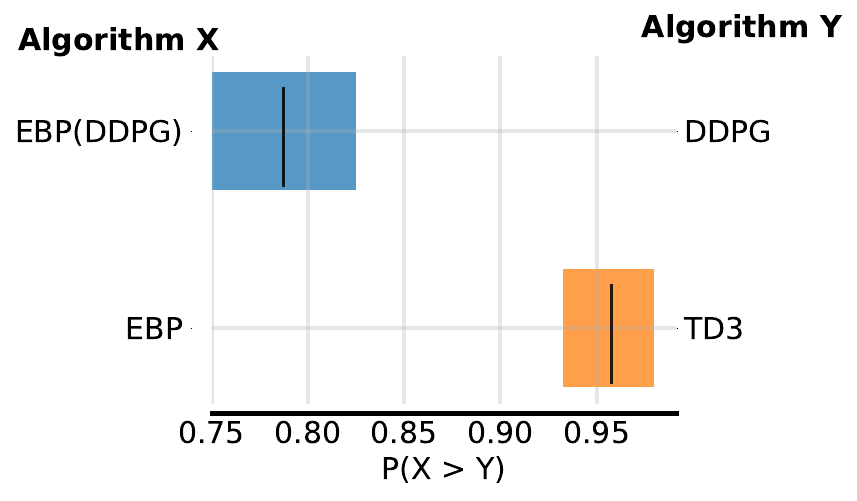}} 
\hfill
  \subfloat[8pt][Online DMControl Evaluation (Full results in Fig.~\ref{fig:Over_results_dmc}).]{
		\includegraphics[height=0.135\textheight]{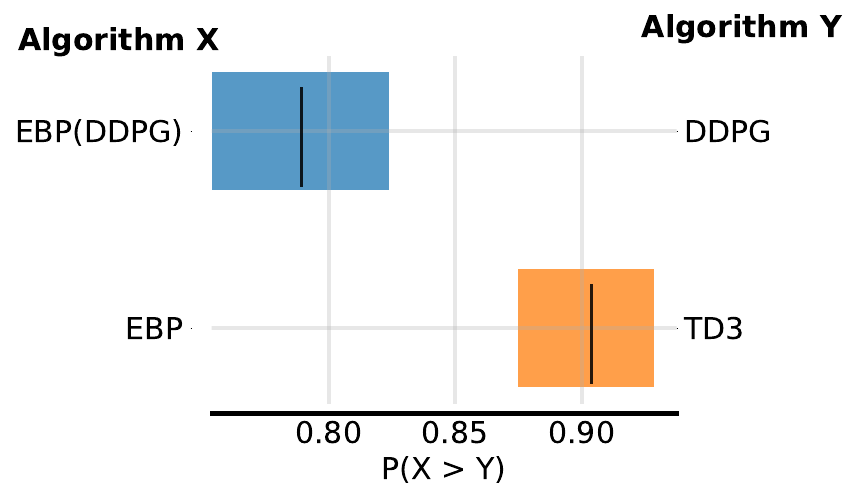} 
        
        }

\caption{RLiable~\cite{agarwal2021deep} analysis allowing us to aggregate results across three continuous control environments and show the probability of improvement for EBP across our empirical evaluation.}

\label{fig:three_envs}
\end{figure*}

\textbf{Meta-analysis of EBP Empirical Evaluation.}
Additionally, we conducted a meta-analysis of our empirical evaluation using the RLiable~\cite{agarwal2021deep} analysis framework. In particular, we report the probability of improvement metric, which quantifies the likelihood that a given algorithm outperforms a reference baseline across tasks and random seeds, as illustrated in Fig.~\ref{fig:three_envs}.

EBP consistently outperforms baseline methods across all three environments, with especially pronounced advantages in the PyBullet and DMControl domains. These results indicate that EBP not only achieves strong peak performance but also maintains stable and reliable behavior across diverse continuous control tasks. Such performance highlights the practical advantages of EBP over existing reinforcement learning algorithms and suggests its strong potential for facilitating the deployment of RL in real-world engineering applications.

\begin{table*}[!htbp]
\centering
\caption{Wall-clock training time and GPU utilization of each algorithm on the HalfCheetah-v3 task, measured over 2M environment steps.}
\begin{tabular}{lcccccccc}
\toprule
Methods        &DDPG&TD3&SAC&BAC&ALH&NNPG&EBP(DDPG)&EBP    \\  
 \midrule
 GPU&450MB& 448MB&491MB&476MB&464MB&2307MB&524MB&540MB\\
Runtime&8.1h& 5.4h & 6.3h&18.3h&6.8h&11.5h&10.4h&11.7h\\
\bottomrule
\end{tabular}
\label{tab:consumer_time}
\end{table*}
\textbf{Runtime and Memory Overhead.}
Finally, to evaluate the runtime cost and GPU memory overhead introduced by the proposed EBP algorithm, we integrate EBP into both TD3 and DDPG and benchmark their computational overhead on the HalfCheetah task. All experiments are conducted on a Linux server equipped with a 56-core Xeon(R) 6133 CPU and a single NVIDIA RTX 3090 GPU. We report the GPU memory consumption and wall-clock training time required for $2$ million environment steps, with the results summarized in Table~\ref{tab:consumer_time}.

The results indicate that, compared to TD3 and DDPG, the training time of EBP and EBP(DDPG) increases by $6.3$ hours and $2.3$ hours, respectively, while GPU memory usage increases by $92$\,MB and $74$\,MB, as shown in Table~\ref{tab:consumer_time}. We attribute this overhead primarily to the additional training and sampling costs introduced by the Q-CVAE module. Moreover, the larger training time increase observed for EBP relative to TD3 stems from the aggressive update strategy adopted by EBP, with the policy update frequency set to $d=1$.

We also observe that NNPG incurs relatively higher GPU memory consumption among all baseline methods, which we attribute to the additional GPU memory required for retrieving nearest neighbors from the offline replay buffer. In contrast, BAC exhibits substantial runtime overhead, likely due to its complex training architecture and multi-branch optimization procedure.

%% file: section/5_Conclusion.tex

\section{Discussion}
\label{appendix_discussion}
These evaluation studies demonstrate that, compared to online policy-prior guidance algorithms, the EBP algorithm leverages a generative expert policy prior as an update anchor to accelerate policy exploitation. The proposed algorithm effectively accelerates policy convergence, thereby leading to superior performance. By incorporating generative expert priors to guide the policy update process, the EBP algorithm effectively addresses the sample inefficiency challenge in online RL.

\textbf{Limitations.}
Despite demonstrating significant performance improvements in experiments, EBP still exhibits some notable limitations.
For instance, as observed in Section~\ref{sec_abla}, the injection of moderate Gaussian reward noise unexpectedly improved performance in the HalfCheetah task. 
This phenomenon can be interpreted through the lens of stochastic uncertainty in Q-networks; however, how to systematically exploit such uncertainty to further enhance performance remains an open question for future investigation.
\section{Conclusions and Future Work}
\label{sec:con_f}
This paper presents a comprehensive study of policy prior guidance algorithms aimed at enhancing sample efficiency. Building upon this analysis, we propose the EBP algorithm, which leverages generative expert policy priors to guide policy updates. By synthesizing high-value actions from a behavior cloning model, EBP provides adaptive and informative supervision throughout the learning process. 
Our extensive empirical evaluation demonstrates that, with a fixed set of hyperparameters, EBP significantly outperforms the current state-of-the-art online RL algorithm (TD3), and consistently surpasses other behavior prior RL algorithms across two robot control and one industrial control environments. In future work, extending the framework to support rapid policy adaptation with limited data under non-stationary or safety-critical environments remains an open and exciting challenge.

%% file: section/6_Appendix.tex
\clearpage
 \appendices
\setcounter{page}{1}

\section{Evaluation Environments and Hyperparameter Setup}

\begin{figure*}[htbp]
\begin{center}
\centerline{\includegraphics[width=0.98\textwidth]{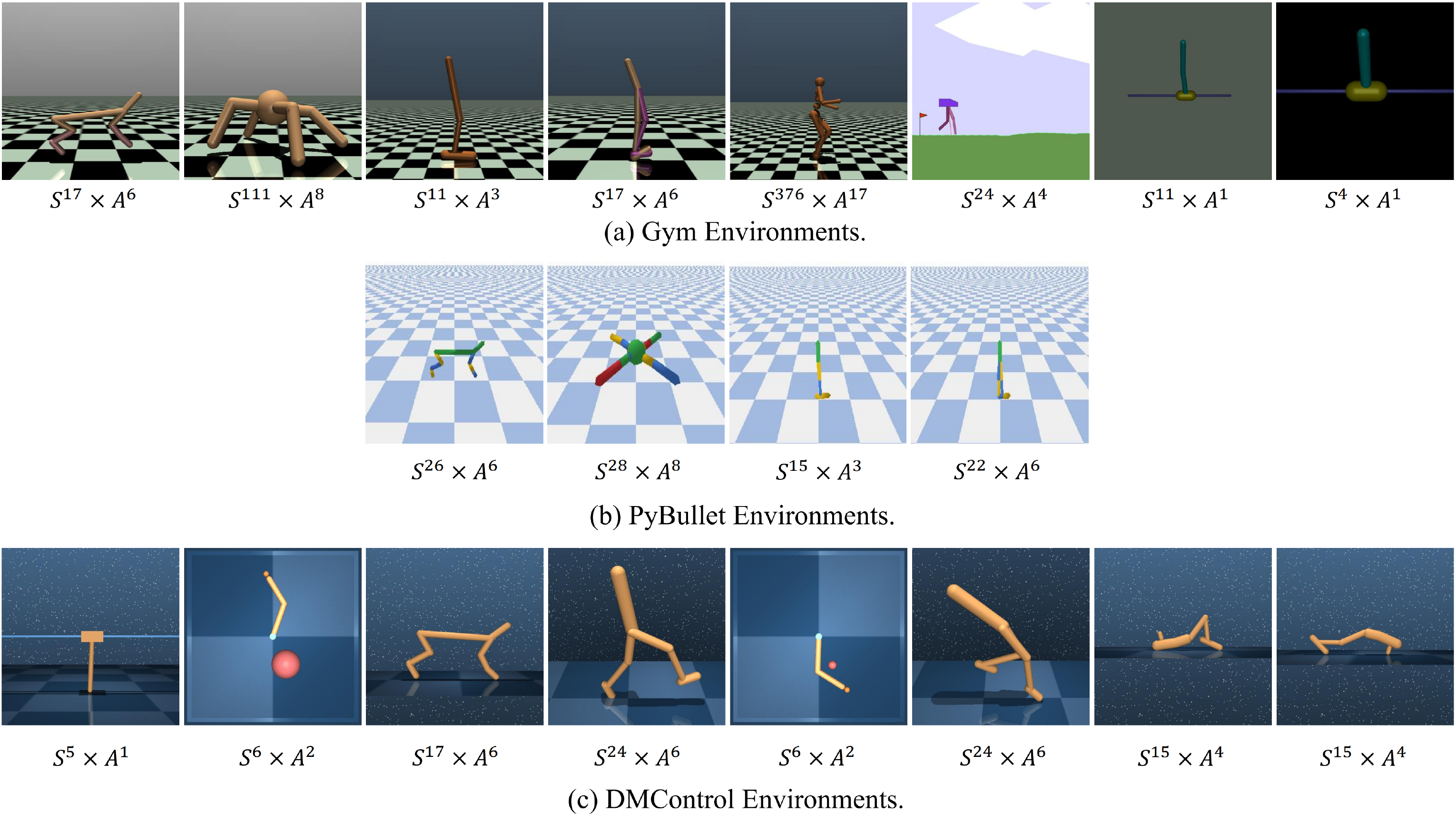}}
\caption{Images for Gym, PyBullet, and DMControl environments used in our experiments.}
\label{task_visual}
\end{center}
\end{figure*}
\subsection{Three Continuous Control Environments}
\label{appendix_datatset}
We evaluated the effectiveness of the EBP algorithm in eight Gym environments, four PyBullet environments, and four DMControl environments, as shown in Fig.~\ref{task_visual}. 
We provided a detailed overview of these environments.

\textbf{Gym.}
Gym is a fast and powerful physics engine that has become a standard platform for simulating complex robotic dynamics and control tasks. It is extensively utilized in the robotics and reinforcement learning communities for training agents to perform tasks such as locomotion and manipulation, and it offers a diverse set of benchmarks for evaluating RL algorithms. In our experiments, we adopt eight challenging tasks from the Gym environments: HalfCheetah, Ant, Hopper, Walker2d, Humanoid, BipedalWalker, InvertedDoublePendulum, and InvertedPendulum.

\textbf{PyBullet.}
PyBullet is an open-source physics engine that supports general-purpose physical simulation as well as high-fidelity modeling for robotics tasks. It provides an efficient control interface for fast simulation of rigid-body dynamics, collision detection, and various control problems. In this paper, we evaluated EBP on four PyBullet benchmark tasks: HalfCheetahBulletEnv, AntBulletEnv, HopperBulletEnv, and Walker2DBulletEnv.

\textbf{DMControl.}
The DMControl suite is a set of continuous control environments mainly designed for robotic manipulation tasks, but it also covers a wide range of industrial control challenges.
In our experiments, we evaluated the proposed algorithm on eight challenging tasks from the DMControl suite: cartpole-swingup, reacher-easy, cheetah-run, walker-walk, reacher-hard, walker-run, hopper-hop, and hopper-stand.

\subsection{Hyperparameter Setup}
\textbf{Q-CVAE Model Structural Design.}  
Our behavior prior model architecture is designed based on the CVAE framework proposed in BCQ~\cite{fujimoto2019off} and MCQ~\cite{lyu2022mildly}. The complete configuration of model parameters is detailed in Table~\ref{table:vaeparams}, which systematically presents the dimensional settings of the encoder, decoder, and the latent space.

\begin{table}[!ht]
  \centering
  \caption{Hyperparameter setup for Q-CVAE}
  \begin{tabular}{ll}
    \toprule
    \multicolumn{2}{c}{\textbf{Q-CVAE Hyperparameters}} \\    
    \midrule  
    Optimizer               & \multicolumn{1}{l}{Adam}            \\
     Q-CVAE learning rate    & \multicolumn{1}{l}{$3\times 10^{-4}$}  \\
    $z$ dimension & $2 \cdot$ action dimension \\
    Hidden activation function & ReLU Layer \\
    Encoder hidden dimension &750\\
    Decoder hidden dimension &750 \\
    Hidden layers &2 \\
    \bottomrule
  \end{tabular}
  \label{table:vaeparams}
\end{table}
\begin{table*}[!htbp]
\setlength\tabcolsep{2.2pt}
\centering
\caption{A complete comparison of hyper-parameter choices between EBP, EBP(DDPG) and six baselines, including DDPG, TD3, SAC, BAC, ALH, and NNPG.}
\begin{tabular}{lcccccccc}
\toprule
\bf{Hyper-parameter} & \bf{DDPG}& \bf{TD3}& \bf{SAC}& \bf{BAC} & \bf{ALH}  &\bf{NNPG}& \bf{EBP(DDPG)}& \bf{EBP}\\
\midrule
Optimizer  & Adam& Adam& Adam&Adam& Adam& Adam& Adam& Adam \\
Critic learning rate & $3 \cdot 10^{-4} $ & $3 \cdot 10^{-4} $& $3 \cdot 10^{-4}$& $3 \cdot 10^{-4}$& $3 \cdot 10^{-4}$& $3 \cdot 10^{-4}$& $3 \cdot 10^{-4}$& $3 \cdot 10^{-4}$  \\
Actor learning rate  & $3 \cdot 10^{-4}$& $3 \cdot 10^{-4} $& $3 \cdot 10^{-4}$&$3 \cdot 10^{-4}$&$3 \cdot 10^{-4}$&$3 \cdot 10^{-4}$&$3 \cdot 10^{-4}$&$3 \cdot 10^{-4}$ \\
Target update rate $\tau$  & $5 \cdot 10^{-3}$& $5 \cdot 10^{-3}$& $5 \cdot 10^{-3}$& $5 \cdot 10^{-3}$ & $5 \cdot 10^{-3}$& $5 \cdot 10^{-3}$& $5 \cdot 10^{-3}$& $5 \cdot 10^{-3}$ \\ 
Batch size  & $256$& $256$& $256$& $512$ & $256$& $256$& $256$& $256$ \\ 
Discount factor & $0.99$ & $0.99$ & $0.99$ & $0.99$ & $0.99$& $0.99$& $0.99$ \\
Number of critics  & 1& 2 &2&2 & 2& 2& 1& 2 \\
Hidden dimension  & $256$& $256$& $256$& $512$ & $256$& $256$& $256$& $256$ \\ 
Start timesteps  & 25e3& 25e3& 1e4& 1e4 & 25e3& 1e4& 1e4& 1e4 \\
Policy update frequency $d$ & 1 &2& 1&1&2&1 \\
 Policy noise $\epsilon$ & $\N(0, 0.2)$& $\N(0, 0.2)$& - &-& $\N(0, 0.2)$ &$\N(0, 0.2)$&$\N(0, 0.2)$&$\N(0, 0.2)$ \\ 
 Noise clip range $c$& $[-0.5,0.5]$& $[-0.5,0.5]$& - & -& $[-0.5,0.5]$ & $[-0.5,0.5]$  & $[-0.5,0.5]$& $[-0.5,0.5]$ \\
   Expected entropy& - & - & -dim(A) &  -dim(A)&-  & -   & - & -  \\
 Q-guided coefficient in CVAE $\alpha$  & -& -& -&-&-&-&0.11&0.11  \\
 Expert policy guidance coefficient $\mu$&-&-&-&-&-&1.0  & 1.0 & 1.0  \\
Exponential $\text{decay\_rate}$ &-&-&-&-&-&0.97&  0.97&  0.97  \\
    Number of policy prior $H$ &-&-&-&-&-&-& 10& 10  \\
    Gradient similarity threshold $m$ &-&-&-&-&-&-& -& 0.05 $\cdot$ action dim.  \\
\bottomrule
\end{tabular}
\label{all_hyper}
\end{table*}
\textbf{Hyperparameter Settings for All Baselines.}
The hyperparameter configurations for EBP and all baseline algorithms used in our experiments are summarized in Table~\ref{all_hyper}.